\documentclass[letterpaper]{article} 
\usepackage{aaai2026}  
\usepackage{times}  
\usepackage{helvet}  
\usepackage{courier}  
\usepackage[hyphens]{url}  
\usepackage{graphicx} 
\usepackage{natbib}  
\usepackage{caption} 
\usepackage{algorithm}
\usepackage{algpseudocode}
\usepackage{amsmath}
\usepackage{amsfonts}
\usepackage{booktabs}
\usepackage{multirow}
\usepackage{xcolor} 
\usepackage{threeparttable}
\usepackage{array}
\nocopyright

\title{D²PPO: Diffusion Policy Policy Optimization with Dispersive Loss}

\author{
    Guowei Zou\textsuperscript{\rm 1,\rm 2},
    Weibing Li\textsuperscript{\rm 1},
    Hejun Wu\textsuperscript{\rm 1,\rm 2},
    Yukun Qian\textsuperscript{\rm 1,\rm 2},
    Yuhang Wang\textsuperscript{\rm 1},
    Haitao Wang\textsuperscript{\rm 1,\rm 2}\thanks{Corresponding author}
}
\affiliations{
    \textsuperscript{\rm 1}School of Computer Science and Engineering, Sun Yat-sen University, Guangzhou, China\\
    \textsuperscript{\rm 2}Guangdong Key Laboratory of Big Data Analysis and Processing, Guangzhou, Guangdong, China\\
    \{zougw, qianyk5, wangyh253, wanght39\}@mail2.sysu.edu.cn, \{liwb53, wuhejun\}@mail.sysu.edu.cn
}

\begin{document}
\maketitle

\begin{abstract}
Diffusion policies excel at robotic manipulation by naturally modeling multimodal action distributions in high-dimensional spaces. Nevertheless, diffusion policies suffer from \textbf{diffusion representation collapse}: semantically similar observations are mapped to indistinguishable features, ultimately impairing their ability to handle subtle but critical variations required for complex robotic manipulation. To address this problem, we propose D²PPO (Diffusion Policy Policy Optimization with Dispersive Loss). D²PPO introduces dispersive loss regularization that combats representation collapse by treating all hidden representations within each batch as negative pairs. D²PPO compels the network to learn discriminative representations of similar observations, thereby enabling the policy to identify subtle yet crucial differences necessary for precise manipulation. In evaluation, we find that early-layer regularization benefits simple tasks, while late-layer regularization sharply enhances performance on complex manipulation tasks. On RoboMimic benchmarks, D²PPO achieves an average improvement of 22.7\% in pre-training and 26.1\% after fine-tuning, setting new SOTA results. In comparison with SOTA, results of real-world experiments on a Franka Emika Panda robot show the excitingly high success rate of our method. The superiority of our method is especially evident in complex tasks. \textbf{\textcolor{blue}{Project page: \url{https://guowei-zou.github.io/d2ppo/}}}
\end{abstract}

\section{Introduction}

Diffusion models have recently emerged as a promising approach for learning robot control policies ~\citep{ddpm2020,ddim2020,dit2023}. Through an iterative denoising mechanism, diffusion policies are able to model complex and multimodal action distributions \citep{lee2025aligning, instructdiffusion2024, diffusionlm2023}, making them well-suited for high-dimensional continuous control tasks~\citep{diffusion_policy2023}.

\begin{figure}[H]
\centering
\includegraphics[width=\columnwidth]{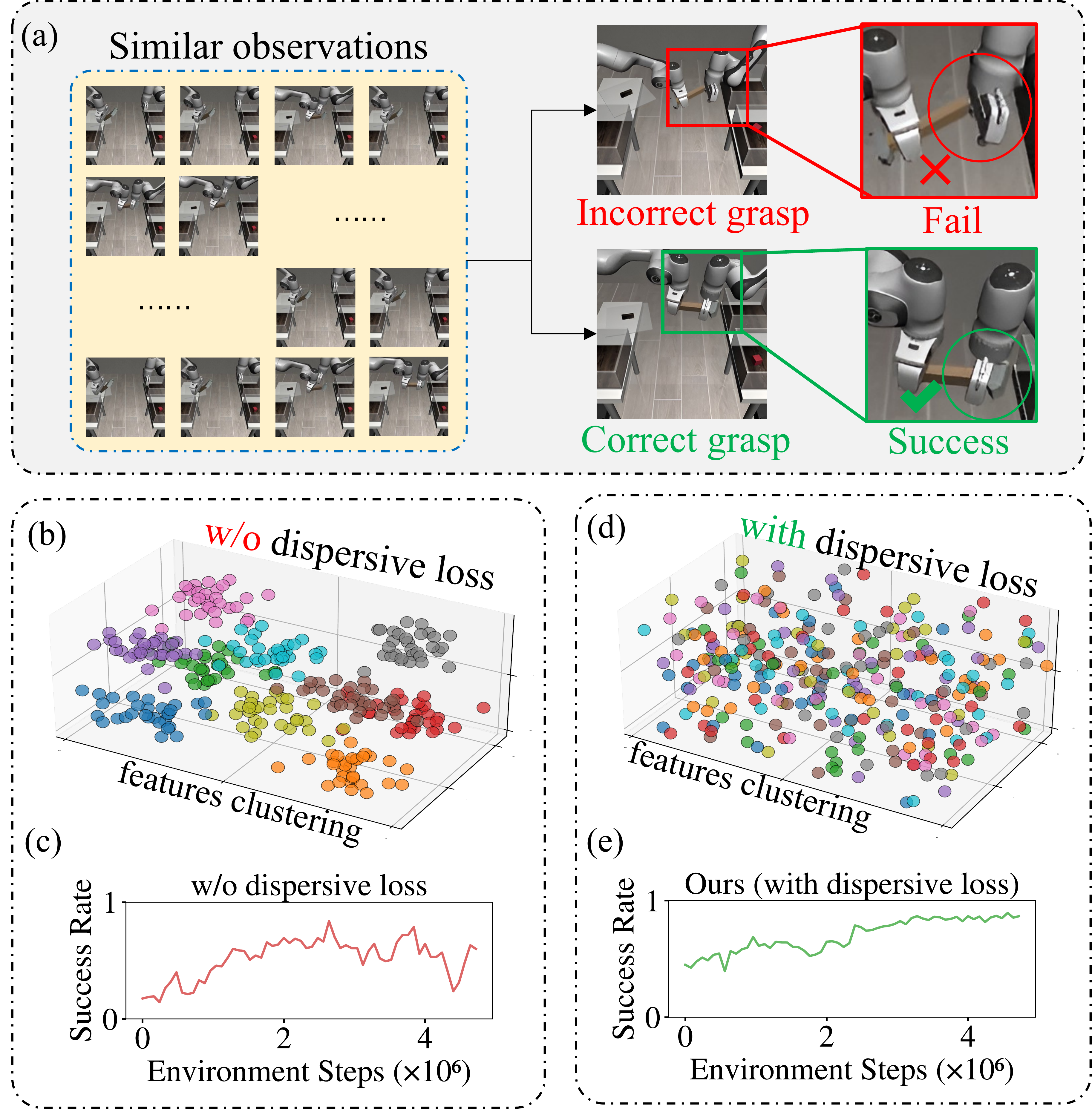}
\caption{
Effect of Dispersive Loss on Representation Quality and Policy Performance.
(a) Representation collapse causes grasping failures when similar observations map to identical features, leading to incorrect decisions between successful (green) and failed (red) attempts. 
(b-c) Without dispersive loss: clustered representations with limited diversity and unstable learning curves.
(d-e) With dispersive loss: well-distributed representations and improved learning with faster convergence.
}
\label{fig:abstract_overview}
\end{figure}

Despite these advantages, we observe that diffusion policies still face significant challenges with low success rates when executing complex manipulation tasks. Consider a robotic arm performing a grasping task where the observations in two scenarios appear highly similar but require distinctly different actions. Standard diffusion policies typically fail to distinguish the subtle yet critical differences between similar observations, consequently generating identical actions that lead to task failure, as shown in Figure~\ref{fig:abstract_overview}(a).

Through detailed analysis of these task failures, we identify the root cause as \textbf{diffusion representation collapse}. This phenomenon arises because diffusion policies rely primarily on reconstruction loss \citep{edm2022, edm2023, consistencymodels2023}. Such loss optimizes for denoising accuracy but neglects the quality and diversity of intermediate feature representations. Consequently, hidden layers produce nearly indistinguishable embeddings for semantically different observations.   As illustrated in Figure~\ref{fig:abstract_overview}(b), clustering analysis of hidden layer features reveals that numerous similar state representations are grouped into nearly identical clusters, indicating poor feature diversity. This representation collapse directly leads to poor performance, as shown in Figure~\ref{fig:abstract_overview}(c). This representation collapse is particularly problematic for robotic manipulation, where subtle observational differences often require substantially different actions for successful task completion. When the learned representations fail to distinguish these nuanced differences, the policy cannot generate the context-sensitive actions required for precise manipulation tasks.





Motivated by advances in computer vision, we try to mitigate this problem through explicit representation regularization. As an explicit form of representation regularization, contrastive representation learning is the most widely used approach for enhancing the discriminability and robustness of feature embeddings~\citep{infonce2018,simclr2020,moco2020, arora2019theoretical, wang2020understanding}. However, traditional contrastive learning typically requires constructing positive-negative sample pairs, additional pre-training stages~\citep{dinov2}, auxiliary model components~\citep{sara2025}, and access to external data for reliable positive pair generation~\citep{repa2024}.

Given these limitations, we propose dispersive loss regularization, a ``\textbf{contrastive loss without positive pairs}" that encourages internal representations to spread out in the hidden space. Specifically, we integrate this dispersive regularization into the standard diffusion loss, maximizing feature dispersion within each batch and thereby enabling the network to distinguish the subtle variations essential for precise manipulation tasks. Importantly, our approach requires no extra pre-training, model parameters, or external data. As illustrated in Figure~\ref{fig:abstract_overview}(d-e), dispersive loss promotes well-separated feature representations and leads to significantly improved learning performance.

Building upon this insight, we propose \textbf{D²PPO (Diffusion Policy Policy Optimization with Dispersive Loss)}, a two-stage training strategy: (1) pre-training with dispersive loss to encourage feature dispersion within each batch; (2) fine-tuning with PPO to maximize task success~\citep{williams1992simple, ppo2017,dppo2024}. This approach combines the generative expressiveness of diffusion models with the goal-directed precision of reinforcement learning, while ensuring that similar observations maintain distinct feature representations crucial for precise manipulation tasks. Detailed theoretical framework and mechanism understanding can be found in Appendix A.

This paper makes three key contributions:
\begin{itemize}
\item Through extensive experiments and analysis, we identify \textbf{diffusion representation collapse} in diffusion policies as the underlying cause of their inability to handle complex manipulation tasks.
\item We propose D²PPO, which addresses diffusion representation collapse by introducing dispersive loss that eliminates the dependency on positive-negative sample pair construction in contrastive learning. We compare three dispersive variants (InfoNCE-L2, InfoNCE-Cosine, Hinge) and analyze their effects across different feature layers.
\item We verify the impact of dispersive loss at different layers on task success rates across varying task complexities, discovering that more challenging tasks benefit increasingly from dispersive regularization. Building on this foundation, we further enhance the trained diffusion policies using the policy gradient algorithm for fine-tuning, achieving improved accuracy with \textbf{real robot validation}.
\end{itemize}

\begin{figure*}[t]
\centering
\includegraphics[width=1\textwidth]{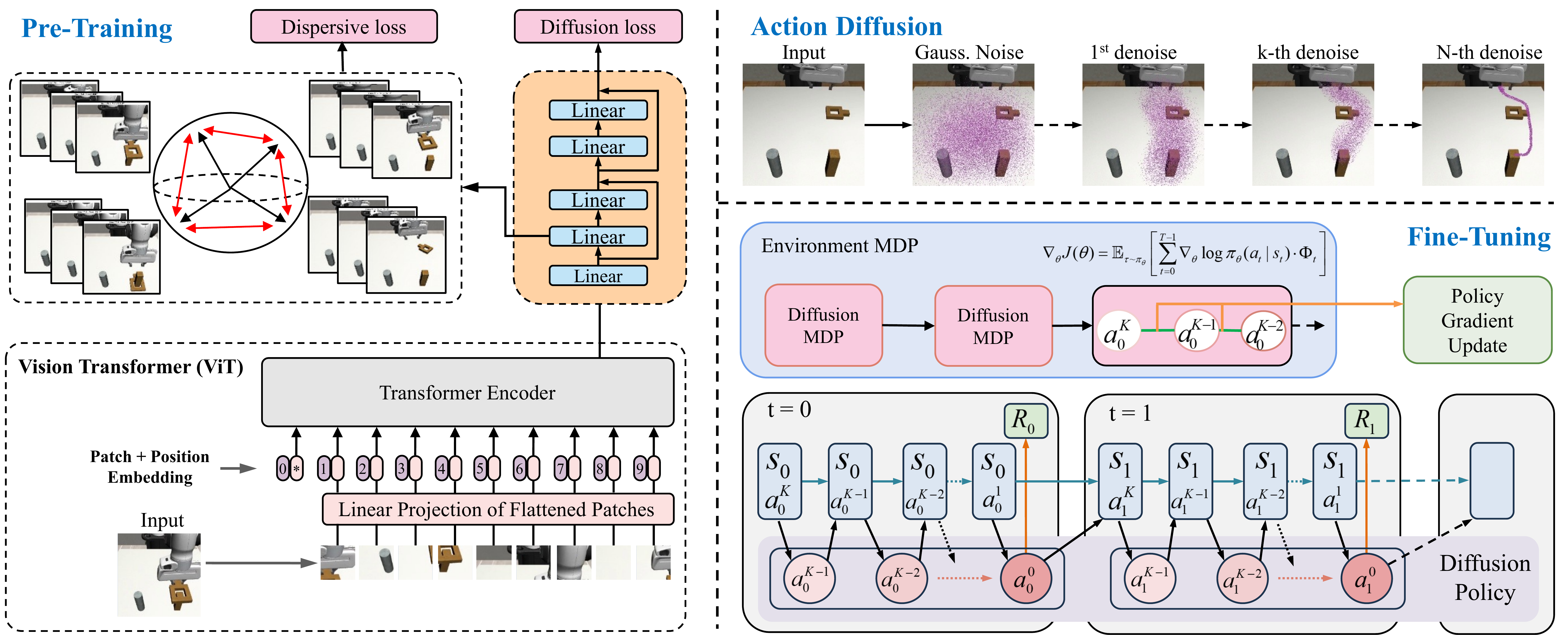}
\caption{D²PPO Framework Overview. The complete two-stage training paradigm: \textbf{Left:} Pre-training stage with Vision Transformer (ViT) feature extraction and dispersive loss regularization to prevent representation collapse; \textbf{Top-right:} Action diffusion process showing iterative denoising from Gaussian noise to final actions; \textbf{Bottom-right:} Fine-tuning stage with policy gradient optimization using two-layer MDP formulation for environment interaction.}
\label{fig:d2ppo_architecture}
\end{figure*}

\section{Related Work}

\textbf{Diffusion Policy for Robot Control.}
Diffusion models revolutionized generative modeling through DDPM~\citep{ddpm2020}, DDIM~\citep{ddim2020}, Latent Diffusion Models~\citep{latentdiffusion2022}, and One step diffusion \citep{shortcutmodels2024}. This paradigm inspired Diffusion Policy~\citep{diffusion_policy2023}, which adapted iterative denoising to robot control by modeling action distributions as denoising processes. Recent extensions include 3D Diffuser Actor~\citep{3ddiffuseractor2024}, 3D Diffusion Policy~\citep{3ddiffusionpolicy2024}, and humanoid manipulation applications~\citep{humanoid3ddp2024}. However, direct application to robotics revealed performance limitations on complex manipulation tasks due to representation collapse, which our work addresses by enhancing diffusion policy representations through dispersive loss regularization to improve manipulation accuracy.

\textbf{Policy Optimization for Diffusion-Based Control.}
Traditional policy gradient methods (TRPO~\citep{trpo2015}, PPO~\citep{gae2015,ppo2017}) required adaptation for diffusion policies' iterative denoising process. Recent approaches include DPPO~\citep{dppo2024} with two-layer MDP formulations, ReinFlow~\citep{reinflow2025, flowql2023} for flow matching with online reinforcement learning, FDPP~\citep{fdpp2025} for human preference integration, and TrajHF~\citep{trajectory_rlhf2025} for human feedback-driven trajectory generation. However, these online fine-tuning methods focus primarily on algorithmic design while neglecting pre-trained diffusion policy representations. Our D²PPO provides a superior starting point for fine-tuning by enhancing representations during pre-training.

\textbf{Representation Learning as the Missing Link.}
Contrastive learning approaches (InfoNCE~\citep{infonce2018}, SimCLR~\citep{simclr2020}, supervised contrastive learning~\citep{supcon2020}) have shown that regularizing learned representations improves generalization. Related approaches in representation regularization include REPA~\cite{repa2024} for alignment with external encoders like DINOv2~\cite{dinov2}, and various contrastive learning extensions building on InfoNCE foundations. Dispersive loss~\citep{dispersive2024} provides an elegant ``contrastive loss without positive pairs" that encourages representational diversity. However, representation methods like dispersive loss have rarely been applied to diffusion policy learning. Our work bridges this insight with diffusion-based robot control, addressing representation collapse that limits complex task performance.

\section{Preliminaries}

\textbf{Diffusion Policy:} Diffusion Policy~\citep{diffusion_policy2023} models the action distribution using a diffusion process. Given an observation $o_t$, the policy generates actions through an iterative denoising process. The forward diffusion process adds Gaussian noise to the action:

\begin{equation}
q(a^1, \ldots, a^K | a^0) = \prod_{k=1}^{K} q(a^k | a^{k-1})
\end{equation}

\noindent where $q(a^k | a^{k-1}) = \mathcal{N}(a^k; \sqrt{1-\beta_k} a^{k-1}, \beta_k \mathrm{I})$, $a^0$ is the original clean action, $a^k$ represents the noisy action at denoising timestep $k$, $K$ is the total number of denoising steps, $\beta_k$ is the noise schedule that controls the amount of noise added at each step, and $\mathrm{I}$ is the identity matrix.

The reverse process learns to predict the noise $\epsilon$ at each timestep:
\begin{equation}
p_\theta(a^{k-1} | a^k, o) = \mathcal{N}(a^{k-1}; \mu_\theta(a^k, k, o), \sigma_k^2 \mathrm{})
\end{equation}

\noindent where $p_\theta$ is the learned reverse process parameterized by $\theta$, $o$ is the observation, $\mu_\theta(a^k, k, o)$ is the predicted mean of the denoising distribution, $\sigma_k^2$ is the variance at timestep $k$, and 
\begin{equation}
\mu_\theta(a^k, k, o) = \frac{1}{\sqrt{\alpha_k}} \left( a^k - \frac{\beta_k}{\sqrt{1-\bar{\alpha}_k}} \epsilon_\theta(a^k, k, o) \right)
\end{equation}
\noindent with $\alpha_k = 1 - \beta_k$, $\bar{\alpha}_k = \prod_{s=1}^k \alpha_s$, and $\epsilon_\theta(a^k, k, o)$ is the neural network that predicts the noise added at timestep $k$.

\noindent \textbf{Dispersive Loss:} The key insight of dispersive loss~\citep{dispersive2024}  is to derive a ``contrastive loss without positive pairs" by reformulating traditional contrastive learning objectives. Traditional contrastive learning optimizes the InfoNCE~\citep{infonce2018} objective, which can be decomposed into two terms:

\begin{equation}
\mathcal{L}_{\text{InfoNCE}} = \frac{\mathcal{D}(z_i, z_i^+)}{\tau} + \log \sum_j \exp\left(-\frac{\mathcal{D}(z_i, z_j)}{\tau}\right)
\end{equation}

\noindent where $z_i$ and $z_i^+$ are positive pairs, $z_j$ are negative samples, $\mathcal{D}(\cdot, \cdot)$ is a distance function, and $\tau$ is the temperature parameter. The first term enforces alignment between positive pairs, while the second term encourages dispersion among all samples. 

The key insight of dispersive loss is to remove the positive pair alignment term entirely, leading to the dispersive objective that focuses solely on representation dispersion:
\begin{equation}
\mathcal{L}_{\text{Disp}} = \log \mathbb{E}_{i,j} \left[ \exp\left(-\frac{\mathcal{D}(z_i, z_j)}{\tau}\right) \right]
\end{equation}

Eq. (5) encourages all representations within a batch to be maximally dispersed in the hidden space, preventing representation collapse and promoting diversity without requiring explicit positive pair construction.

\section{Method}
\textbf{D²PPO: Enhanced Diffusion Policy Policy Optimization using Dispersive Loss.}
D²PPO addresses representation collapse in diffusion policies through dispersive loss regularization. D²PPO adopts a two-stage training paradigm: dispersive pre-training followed by policy gradient optimization, as illustrated in Figure~\ref{fig:d2ppo_architecture}. Complete mathematical derivations are provided in Appendix B. Algorithm pseudocode is provided in Appendix C.

\subsubsection{Stage 1: Enhanced Pre-training with Dispersive Loss.} We augment the standard diffusion loss with dispersive regularization. Our approach employs a Vision Transformer encoder (ViT) \citep{transformer2017, vit2021} for visual feature extraction and applies regularization to selected intermediate layers of the MLP denoising network. We define our pre-training objective as:
\begin{equation}
\mathcal{L}_{\text{D}^2\text{PPO}}^{\text{pre-train}} = \mathcal{L}_{\text{diff}} + \lambda \mathcal{L}_{\text{disp}}
\end{equation}

This combined objective serves two purposes: the diffusion loss $\mathcal{L}_{\text{diff}}$ ensures accurate noise prediction for proper denoising, while the dispersive loss $\mathcal{L}_{\text{disp}}$ with weight $\lambda$ encourages representational diversity in the learned features.

The dispersive loss is computed by averaging over all denoising timesteps:
\begin{equation}
\mathcal{L}_{\text{disp}} = \frac{1}{K} \sum_{k=1}^{K} \mathcal{L}_{\text{disp}}^{\text{variant}}(\mathbf{H}_{k})
\end{equation}

\noindent where $K$ denotes the total number of denoising steps in the diffusion process, $k$ is the current denoising timestep index, and $\mathcal{L}_{\text{disp}}^{\text{variant}}(\mathbf{H}_{k})$ represents the specific dispersive loss variant (InfoNCE-L2, InfoNCE-Cosine, or Hinge) computed at timestep $k$. The notation $\mathbf{H}_{k} = \{h_{i,k}\}_{i=1}^B$ denotes the collection of intermediate feature representations from all $B$ samples in the batch at timestep $k$.

We implement three main variants of dispersive loss, each derived from different contrastive learning methods by removing the positive alignment term:

\noindent \textbf{1. InfoNCE-based Dispersive Loss with L2 Distance:}
\begin{equation}
\mathcal{L}_{\text{disp}}^{\text{InfoNCE-L2}} = \log \mathbb{E}_{i,j} \left[ \exp\left(-\frac{||h_i - h_j||_2^2}{\tau}\right) \right]
\end{equation}

Eq. (8) uses squared $\ell_2$ distance $\mathcal{D}(h_i, h_j) = ||h_i - h_j||_2^2$, which measures Euclidean distance in the representation space. This formulation encourages dispersion based on geometric distance between feature vectors.

\noindent \textbf{2. InfoNCE-based Dispersive Loss with Cosine Distance:}
\begin{equation}
\mathcal{L}_{\text{disp}}^{\text{InfoNCE-Cos}} = \log \mathbb{E}_{i,j} \left[ \exp\left(-\frac{1 - \frac{h_i^T h_j}{||h_i||_2 \cdot ||h_j||_2}}{\tau}\right) \right]
\end{equation}

Eq. (9) uses cosine dissimilarity $\mathcal{D}(h_i, h_j) = 1 - {h_i^T h_j}/{||h_i||_2 \cdot ||h_j||_2}$, which captures angular differences between normalized representations. This formulation is scale-invariant and focuses on directional diversity.

\noindent \textbf{3. Hinge Loss-based Dispersive Loss:}
\begin{equation}
\mathcal{L}_{\text{disp}}^{\text{Hinge}} = \mathbb{E}_{i,j} \left[ \max(0, \epsilon - \mathcal{D}(h_i, h_j))^2 \right]
\end{equation}

Eq. (10) is derived from hinge loss by removing the positive pair term, focusing purely on enforcing a minimum margin $\epsilon$ between all representation pairs. This formulation directly penalizes representations that are closer than the margin threshold, providing explicit control over the minimum dispersion distance.

In our experiments, we evaluate all three variants across different network layers to determine optimal configurations for each task complexity level.

\subsubsection{Stage 2: Dispersive loss-augmented diffusion policy optimization.} In the optimization stage, D²PPO leverages the enhanced representations learned during pre-training to optimize diffusion policies through reinforcement learning. Our approach strategically focuses this stage on reward maximization while preserving the representational structure established during pre-training.

\textbf{Overall Objective:} We aim to optimize the diffusion policy $\pi_\theta$ to maximize expected return:
\begin{equation}
J(\theta) = \mathbb{E}_{\tau \sim \pi_\theta}[R(\tau)]
\end{equation}

For diffusion policies, the action probability involves the entire denoising chain from $a_t^K$ (pure noise) to $a_t^0$ (final action):
\begin{equation}
\pi_\theta(a_t^0 | s_t) = \int p(a_t^K) \prod_{k=1}^{K} p_\theta(a_t^{k-1} | a_t^k, s_t) da_t^{1:K}
\end{equation}

\noindent where the diffusion policy is a multi-step latent policy where we cannot directly obtain $\pi(a|s)$ as in standard policies, but must consider the joint probability across the entire denoising chain.

Using the chain rule, the log probability gradient becomes:
\begin{equation}
\nabla_\theta \log \pi_\theta(a_t^0 | s_t) = \sum_{k=1}^{K} \nabla_\theta \log p_\theta(a_t^{k-1} | a_t^k, s_t)
\end{equation}

Since the previous step involves multi-step generation, we can only compute gradients for each conditional probability step and accumulate them using the chain rule. This step crucially transforms the entire policy gradient into differentiable losses over individual denoising steps, enabling gradient-based optimization.

To accelerate training, we employ importance sampling since computing gradients for all $k = 1, \ldots, K$ steps is computationally expensive:
\begin{equation}
\nabla_\theta J(\theta) \approx \frac{1}{|\mathcal{S}|} \sum_{k \in \mathcal{S}} \frac{K}{p(k)} \nabla_\theta \log p_\theta(a_t^{k-1} | a_t^k, s_t) \cdot \hat{A}_t^{(k)}
\end{equation}

\noindent where $\mathcal{S}$ denotes the sampled subset of denoising steps, $|\mathcal{S}|$ is the subset size, $p(k)$ is the sampling probability for step $k$, and $\hat{A}_t^{(k)}$ is the step-conditioned advantage estimate. This formulation enables efficient policy gradient computation while maintaining the enhanced representational structure from pre-training.

\begin{figure*}[t]
\centering
\includegraphics[width=\textwidth]{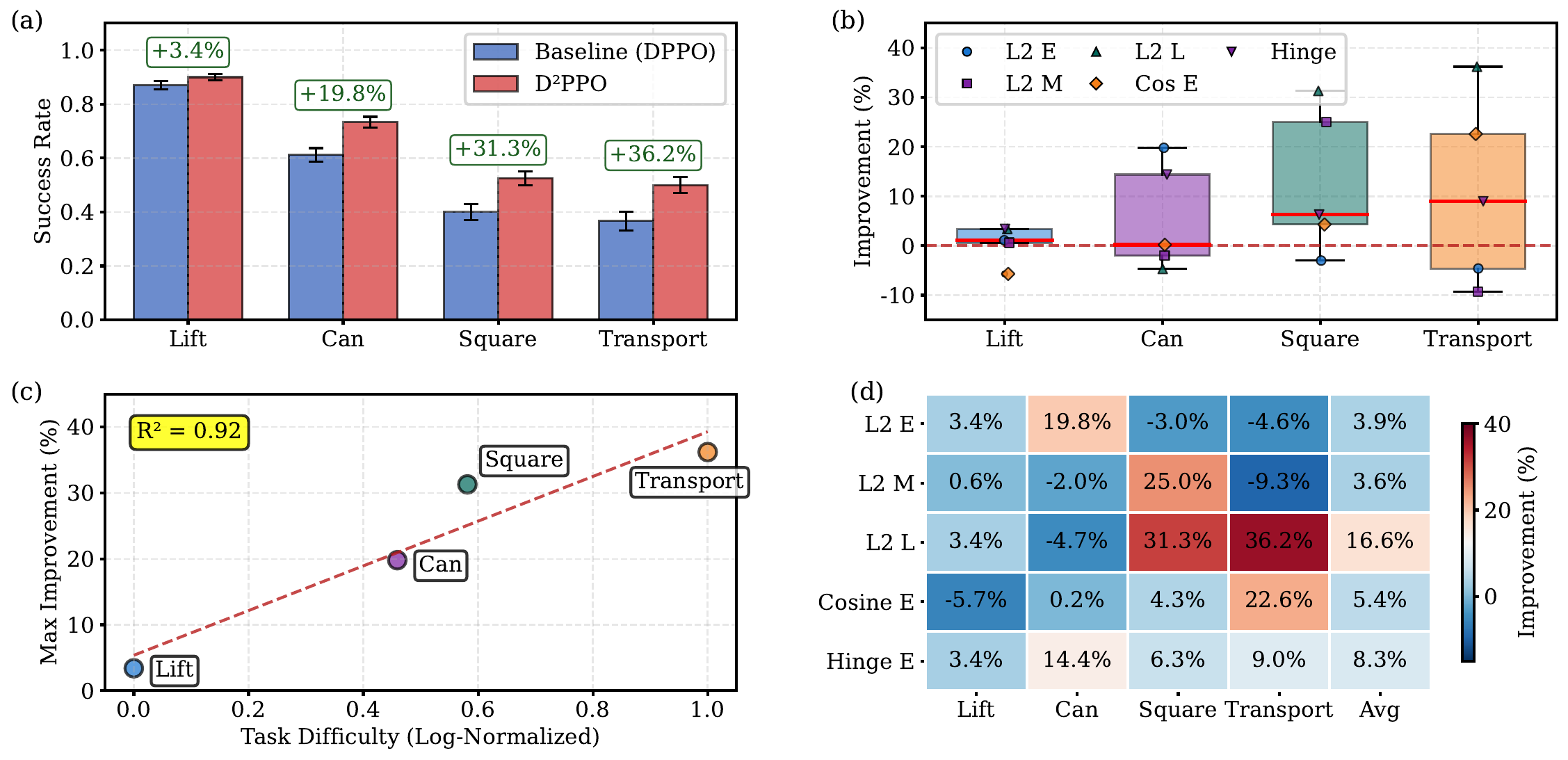}
\caption{Comprehensive pre-training experimental results using D²PPO with dispersive loss across four robotic manipulation tasks. (a) Performance comparison showing baseline (DPPO~\citep{dppo2024}) versus D²PPO success rates with error bars, demonstrating consistent improvements across all tasks. (b) Distribution of improvement rates across five dispersive loss variants (InfoNCE L2 Early/Mid/Late, InfoNCE Cosine Early, Hinge Early). (c) Task difficulty correlation analysis showing the relationship between log-normalized task complexity and maximum improvement rates. (d) Method suitability matrix heatmap displaying improvement percentages for each dispersive loss variant across tasks, with color intensity indicating effectiveness.}
\label{fig:comprehensive_four_subplots}
\end{figure*}

\section{Experiments}

We evaluate D²PPO through a comprehensive three-stage experimental evaluation: (1) pre-training experiments that validate dispersive loss effectiveness, (2) fine-tuning experiments that demonstrate superior performance compared to existing algorithms, and (3) real robot experiments that showcase practical deployment capabilities. We conduct experiments on four representative tasks from the robomimic benchmark~\citep{robomimic2022, mujoco2012}: Lift, Can, Square, and Transport, spanning different manipulation complexities. Detailed experimental configurations, datasets, and supplementary experiments are provided in Appendix D.

\subsection{Pre-training Experiments with Dispersive Loss}

We first evaluate the effectiveness of dispersive loss regularization during the pre-training stage, examining multiple variants and configurations to validate the impact on representation learning and policy performance.

The four tasks represent increasing complexity levels: \textbf{Lift} (basic object manipulation), \textbf{Can} (cylindrical object grasping), \textbf{Square} (precise peg-in-hole placement), and \textbf{Transport} (multi-object coordination). We define task difficulty using log-normalized execution steps, providing objective complexity quantification as shown in Table~\ref{tab:task_difficulty}:

\begin{table}[ht]
\centering
\small
\begin{tabular}{lccc}
\toprule
Task & Steps & Difficulty & Complexity \\
\midrule
\textbf{Lift} & 108 & 0.000 & Easiest \\
\textbf{Can} & 224 & 0.459 & Moderate \\
\textbf{Square} & 272 & 0.581 & High \\
\textbf{Transport} & 529 & 1.000 & Highest \\
\bottomrule
\end{tabular}
\caption{Task difficulty hierarchy based on average execution steps and log-normalized difficulty scores.}
\label{tab:task_difficulty}
\end{table}

\begin{figure}[ht]
\centering
\includegraphics[width=0.9\columnwidth]{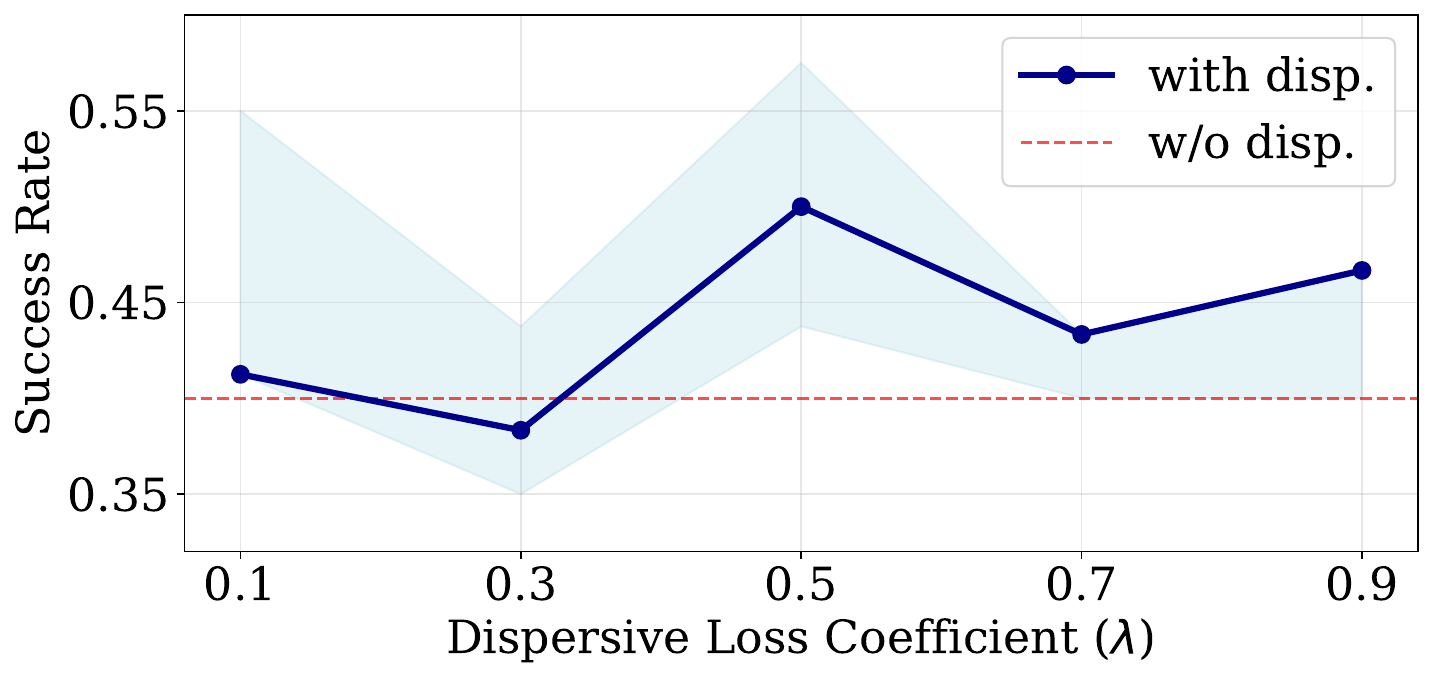}
\caption{Effect of dispersive loss coefficient $\lambda$ on Square task performance.}
\label{fig:lambda_effect_square}
\end{figure}

\begin{figure*}[t]
\centering
\includegraphics[width=\textwidth]{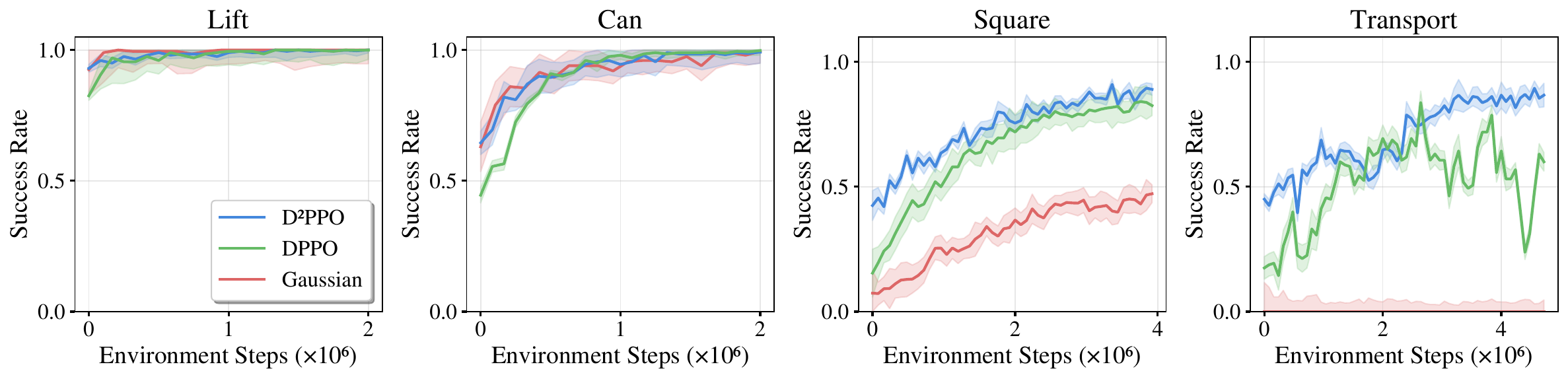}
\caption{Policy gradient fine-tuning results across four robotic manipulation tasks. The learning curves demonstrate that D²PPO consistently achieves superior sample efficiency and final performance compared to baseline DPPO and Gaussian policies.}
\label{fig:robomimic_finetuning}
\end{figure*}

\begin{table*}[t]
\centering
\small
\caption{Performance comparison on robomimic tasks. \textbf{Bold} indicates best performance, \underline{underline} indicates second-best.}
\begin{tabular}{l*{4}{>{\centering\arraybackslash}p{1.2cm}}*{4}{c}}
\toprule
\multirow{2}{*}{Method} & \multicolumn{4}{c}{Success Rate} & \multirow{2}{*}{Avg.} & \multirow{2}{*}{Median} & \multirow{2}{*}{Std.} & \multirow{2}{*}{Rank} \\
\cmidrule(lr){2-5}
& Lift & Can & Square & Transport & & & & \\
\midrule
LSTM-GMM~\citep{robomimic2022} & 0.93 & 0.81 & 0.59 & 0.20 & 0.63 & 0.70 & 0.32 & 7 \\
IBC~\citep{ibc2022} & 0.15 & 0.01 & 0.00 & 0.00 & 0.04 & 0.01 & 0.07 & 9 \\
BET~\citep{bet2022} & \textbf{1.00} & 0.90 & 0.43 & 0.06 & 0.60 & 0.67 & 0.41 & 8 \\
DiffusionPolicy-C~\citep{diffusion_policy2023} & \underline{0.97} & \underline{0.96} & 0.82 & 0.46 & 0.80 & 0.89 & 0.23 & 4 \\
DiffusionPolicy-T~\citep{diffusion_policy2023} & \textbf{1.00} & 0.94 & 0.81 & 0.35 & 0.78 & 0.88 & 0.29 & 5 \\
DPPO~\citep{dppo2024} & \textbf{1.00} & \textbf{1.00} & \underline{0.83} & \underline{0.60} & \underline{0.86} & \underline{0.92} & 0.18 & \underline{2} \\
MTIL (10-step)~\citep{mtil2025} & 0.94 & 0.83 & 0.61 & 0.22 & 0.65 & 0.72 & 0.31 & 6 \\
MTIL (Full History)~\citep{mtil2025} & \textbf{1.00} & \underline{0.96} & \underline{0.83} & 0.48 & 0.82 & 0.90 & 0.23 & 3 \\
\midrule
\textbf{D²PPO (Ours)} & \textbf{1.00} & \textbf{1.00} & \textbf{0.89} & \textbf{0.87} & \textbf{0.94} & \textbf{0.95} & \textbf{0.06} & \textbf{1} \\
\bottomrule
\end{tabular}
\label{tab:robomimic_mh_core}
\end{table*}

Figure~\ref{fig:comprehensive_four_subplots} presents comprehensive pre-training experimental results across all five dispersive loss variants and four robotic manipulation tasks. Figure~\ref{fig:comprehensive_four_subplots}(a) shows D²PPO achieving consistent improvements across all tasks, with enhancement rates ranging from +3.4\% (Lift) to +36.2\% (Transport). Figure~\ref{fig:comprehensive_four_subplots}(b) reveals the distribution of improvement rates across five dispersive loss variants, with distinct markers for each algorithm type. 
Figure~\ref{fig:comprehensive_four_subplots}(c) demonstrates a strong positive correlation (R² = 0.92) between log-normalized task difficulty and maximum improvement rates, validating our hypothesis that representation quality becomes increasingly critical for complex robotic tasks. Figure~\ref{fig:comprehensive_four_subplots}(d) provides empirical guidance for layer selection through a heatmap showing dispersive loss effectiveness at different network layers, establishing that simple tasks achieve optimal performance with early-layer application while complex tasks require late-layer regularization. 

These comprehensive pre-training results establish that D²PPO with dispersive loss achieves superior performance compared to DPPO, with an average improvement of 22.7\% across all tasks.

To further investigate the impact of the dispersive loss coefficient $\lambda$, we conduct detailed ablation studies on the Square task, which requires precise spatial coordination and is particularly sensitive to representation quality. Figure~\ref{fig:lambda_effect_square} reveals a non-monotonic relationship: performance peaks at $\lambda=0.5$ (14.3\% improvement over baseline), while $\lambda=0.3$ degrades below baseline performance. This suggests insufficient regularization fails to encourage adequate representation diversity, while excessive values ($\lambda>0.5$) may interfere with task-relevant learning, highlighting the importance of balanced dispersive regularization.

\begin{figure*}[t]
\centering
\includegraphics[width=\textwidth]{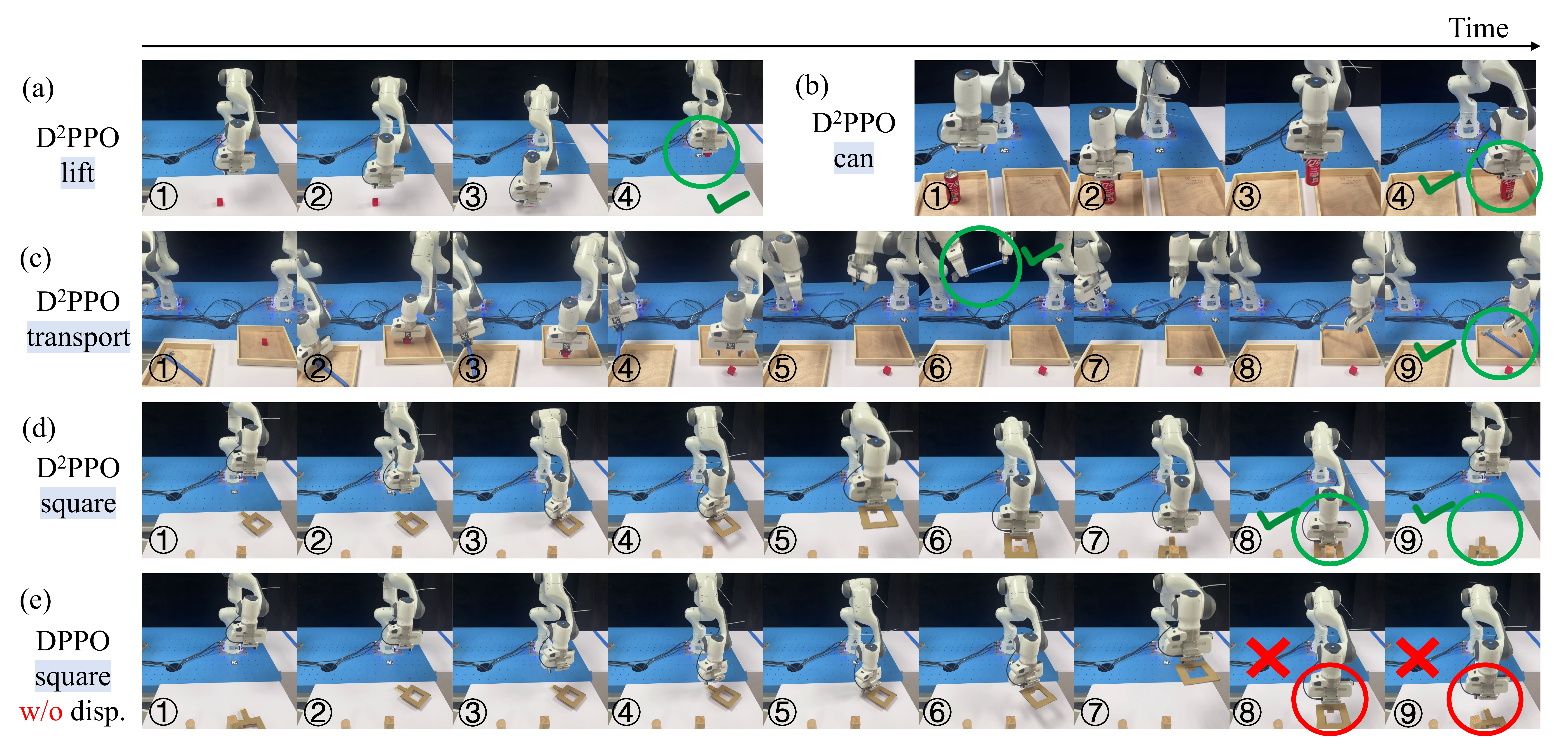}
\caption{Real-world deployment results on RoboMimic tasks using D²PPO. Each row shows the complete task execution trajectory over time. (a-d) D²PPO with dispersive loss successfully completes all tasks: (a) Lift task with successful block grasping, (b) Can task with successful cylindrical object manipulation, (c) Transport task with successful object delivery to target area, and (d) Square task with precise peg-in-hole insertion (green checkmarks indicate successful completion with green circles highlighting key successful actions). (e) Baseline DPPO without dispersive loss fails in the Square task during final placement phases (red crosses and circles indicate failure points), demonstrating the effectiveness of dispersive representation learning for precise manipulation.}
\label{fig:sim_to_real}
\end{figure*}

\subsection{Fine-tuning Experiments}

As the second phase of our evaluation, we validate the complete D²PPO method through comprehensive policy gradient fine-tuning experiments, demonstrating how enhanced representations translate into superior reinforcement learning performance and comparing against existing SOTA algorithms. For each task, we select the best-performing pre-trained model weights from our dispersive loss variants and use them as initialization for policy gradient fine-tuning.

Figure~\ref{fig:robomimic_finetuning} presents the fine-tuning learning curves across all four tasks, revealing several critical insights: \textbf{(1) Enhanced Sample Efficiency:} D²PPO demonstrates consistently faster convergence across almost all tasks, requiring significantly fewer environment interactions to achieve comparable performance levels. \textbf{(2) Superior Asymptotic Performance:} The final performance levels achieved by D²PPO consistently exceed baseline methods, with improvements being most dramatic in challenging manipulation scenarios. \textbf{(3) Training Stability:} D²PPO exhibits more stable learning dynamics with reduced variance, indicating that the enhanced representation foundation improves optimization robustness.

D²PPO consistently outperforms both DPPO and Gaussian-based algorithms across all four tasks. Notably, in complex manipulation scenarios, D²PPO achieves substantial improvements: Square task performance increases from 47\% (Gaussian) $\rightarrow$ 83\% (DPPO) $\rightarrow$ 89\% (Ours), while Transport task shows progression from 0\% (Gaussian) $\rightarrow$ 60\% (DPPO) $\rightarrow$ 87\% (Ours), demonstrating the critical importance of diffusion-based representations with dispersive regularization for complex manipulation.

Table~\ref{tab:robomimic_mh_core} reveals several critical insights into the effectiveness of our two-stage D²PPO approach. Most significantly, D²PPO achieves SOTA performance across
all tasks, ranking first overall with an average success rate of 0.94, representing an average improvement of 26.1\% across the four tasks. These fine-tuning results provide definitive validation that our two-stage training paradigm successfully translates representation enhancements into practical performance improvements. Dispersive pre-training provides enhanced representations that improve policy gradient optimization, enabling more efficient learning and better performance across robotic manipulation tasks.

\subsection{Real Robot Experiments}

As the final validation of our approach, we demonstrate the practical deployment capability of D²PPO through real robot experiments on the Franka Emika Panda robot. These experiments are conducted after completing the two-stage training process (dispersive pre-training followed by PPO fine-tuning) and showcase comprehensive evaluation across multiple RoboMimic tasks with direct comparison to baseline methods. On the most challenging transport task, we observe performance progression: 0\% (Gaussian) $\rightarrow$ 45\% (DPPO) $\rightarrow$ 70\% (Ours). Quantitative experimental results are detailed in Appendix D.

Figure 6 presents qualitative results from real robot experiments. These results reveal several critical insights: \textbf{(1) Task Completion Success:} D²PPO with dispersive loss successfully completes all four benchmark tasks, with each trajectory showing complete execution sequences from initial approach to final success confirmation. \textbf{(2) Precision in Complex Tasks:} The Square task, requiring precise peg-in-hole insertion, demonstrates D²PPO's ability to handle fine-grained manipulation where spatial accuracy is critical. \textbf{(3) Baseline Comparison:} DPPO without dispersive loss fails to complete the Square task, with clear failure points visible in the final placement phases, highlighting the critical importance of dispersive representation learning for precise manipulation scenarios.

\section{Conclusion}

This paper addresses representation collapse in diffusion-based policy learning, where similar observations lead to indistinguishable features, causing failures in complex manipulation tasks. We introduce D²PPO, a two-stage framework applying dispersive loss regularization during pre-training, followed by policy gradient optimization for fine-tuning.

Our contributions include: (1) systematic analysis identifying representation collapse as the underlying cause of poor performance in complex tasks; (2) D²PPO method that combats representation collapse through dispersive regularization; (3) comprehensive validation demonstrating state-of-the-art performance with 94\% average success rate, achieving 22.7\% improvement in pre-training and 26.1\% improvement after fine-tuning compared to baseline methods.

Experimental results reveal that representation quality becomes increasingly critical for complex tasks, with D²PPO showing universal effectiveness across varying task complexities. D²PPO requires no additional parameters and provides plug-and-play integration with existing diffusion policies. Real robot validation confirms practical deployment effectiveness, establishing dispersive loss as a powerful tool for enhancing diffusion-based robotic policies and opening new directions for representation-regularized policy learning. While our current evaluation focuses on manipulation tasks, future work could explore dispersive regularization in other robotic domains.

\bibliography{d2ppo-arxiv}

\clearpage
\onecolumn

\section{\raggedright\Large Appendix A: Theoretical Framework and Mechanistic Understanding}

The effectiveness of dispersive loss can be understood through representation capacity and policy expressiveness in diffusion-based learning systems. In diffusion policies, intermediate representations form hierarchical feature spaces encoding increasingly refined action information through iterative denoising processes. When these representations lack diversity due to reconstruction-focused training objectives, policies become susceptible to representational mode collapse, learning limited behavioral patterns despite 
the rich expressiveness potential inherent in diffusion model architectures.

The observed correlation between task complexity and dispersive loss effectiveness (R² = 0.92) can be understood through information-theoretic analysis of robotic manipulation requirements. Complex manipulation tasks require policies to distinguish between exponentially larger numbers of behavioral modes and environmental configurations compared to simpler tasks. From a mutual information perspective, complex tasks require higher mutual information between observations and optimal actions to achieve successful task completion, necessitating more diverse and informative internal representations that can capture subtle but critical state differences.

The core theoretical value of dispersive loss in robotic manipulation lies in addressing extreme precision requirements characteristic of physical manipulation tasks. In robotic scenarios, environmental states appearing nearly identical to standard feature extractors may require completely different optimal actions for successful task completion. For example, during precision grasping operations, two robot configurations differing by mere millimeters in end-effector position may necessitate opposite motion directions to achieve successful object manipulation. Traditional representation learning approaches that allow similar states to cluster in feature space lead to averaged action predictions that systematically fail in practice. Dispersive loss addresses this fundamental limitation by forcing networks to learn distinct representations for all observed states, enabling precise state-action mappings essential for fine motor control and manipulation success.

\section{\raggedright\Large Appendix B: Mathematical Derivations for D²PPO}

\subsection{Notation and Variable Definitions}

This section provides a comprehensive reference for all mathematical symbols and variables used throughout the D²PPO framework, covering self-supervised contrastive learning, diffusion policy supervised learning, and PPO reinforcement learning components.

This work uses a consistent notation system where $t$ denotes reinforcement learning timesteps and $k$ denotes diffusion denoising timesteps.

\begin{table}[ht]
\centering
\small
\begin{tabular}{cl}
\toprule
\textbf{Symbol} & \textbf{Description} \\
\midrule
\multicolumn{2}{c}{\textit{General Notation}} \\
$\theta$ & Neural network parameters \\
$\phi$ & Value function parameters \\
$B$ & Batch size \\
$N$ & Total number of samples \\
$I$ & Identity matrix \\
$\mathbb{E}[\cdot]$ & Expectation operator \\
$\nabla_\theta$ & Gradient with respect to parameters $\theta$ \\
$||\cdot||_2$ & L2 (Euclidean) norm \\
$\mathcal{N}(\mu, \sigma^2)$ & Gaussian distribution with mean $\mu$ and variance $\sigma^2$ \\
\midrule
\multicolumn{2}{c}{\textit{Environment and Task Variables}} \\
$s_t, o_t$ & State/observation at time $t$ \\
$a_t$ & Action at time $t$ \\
$r_t$ & Reward at time $t$ \\
$R_t$ & Return (cumulative reward) from time $t$ \\
$\tau$ & Trajectory $\{s_0, a_0, r_0, s_1, a_1, r_1, \ldots\}$ \\
$V_\phi(s_t)$ & Value function estimate at state $s_t$ \\
$\hat{A}_t$ & Advantage estimate at time $t$ \\
$\hat{A}_t^{(k)}$ & Step-conditioned advantage estimate for denoising step $k$ \\
\midrule
\multicolumn{2}{c}{\textit{Diffusion Process Variables}} \\
$a_0$ & Original clean action (target) \\
$a^k$ & Noisy action at denoising timestep $k$ \\
$a_t^k$ & Action at RL timestep $t$, denoising step $k$ \\
$a^K$ & Pure noise (initial diffusion state) \\
$K$ & Total number of denoising timesteps \\
$k$ & Current denoising timestep index \\
$\beta_k$ & Noise schedule parameter at timestep $k$ \\
$\alpha_k$ & Noise retention parameter, $\alpha_k = 1 - \beta_k$ \\
$\bar{\alpha}_k$ & Cumulative noise parameter, $\bar{\alpha}_k = \prod_{i=1}^k \alpha_i$ \\
$\sigma_k^2$ & Variance of denoising distribution at timestep $k$ \\
$\epsilon_\theta$ & Neural network predicting added noise \\
$\mu_\theta$ & Predicted mean of denoising distribution \\
\bottomrule
\end{tabular}
\caption{Notation and variable definitions for D²PPO framework - Part I}
\label{tab:notation_part1}
\end{table}

\begin{table}[ht]
\centering
\small
\begin{tabular}{cl}
\toprule
\textbf{Symbol} & \textbf{Description} \\
\midrule
\multicolumn{2}{c}{\textit{Policy and Probability Variables}} \\
$\pi_\theta$ & Policy parameterized by $\theta$ \\
$\pi_{\theta_{old}}$ & Old policy (for PPO importance sampling) \\
$p_\theta(a^{k-1}|a^k, o)$ & Denoising transition probability \\
$p_\theta(a_t^{k-1}|a_t^k, s_t)$ & RL-conditioned denoising transition \\
$q(a^k|a^{k-1})$ & Forward diffusion transition \\
$r_t^{(k)}(\theta)$ & Probability ratio for PPO at denoising step $k$ \\
$J(\theta)$ & Policy objective function \\
$p(k)$ & Sampling probability for timestep $k$ \\
$\mathcal{S}$ & Sampled subset of denoising timesteps \\
$|\mathcal{S}|$ & Size of sampled timestep subset \\
\midrule
\multicolumn{2}{c}{\textit{Representation Learning Variables}} \\
$h_i, z_i$ & Feature representation of sample $i$ \\
$h_{i,k}$ & Feature representation of sample $i$ at denoising step $k$ \\
$\mathbf{H}_k$ & Collection of all features at timestep $k$, $\{h_{i,k}\}_{i=1}^B$ \\
$z_i^+$ & Positive pair for sample $i$ (traditional contrastive learning) \\
$z_j$ & Negative sample $j$ \\
$w_{ij}$ & Attention weight between representations $i$ and $j$ \\
\midrule
\multicolumn{2}{c}{\textit{Loss Function Variables}} \\
$\mathcal{L}_{InfoNCE}$ & InfoNCE contrastive loss \\
$\mathcal{L}_{Disp}$ & Dispersive loss (general) \\
$\mathcal{L}_{\text{disp}}^{\text{L2}}$ & Dispersive loss with L2 distance \\
$\mathcal{L}_{\text{disp}}^{\text{cos}}$ & Dispersive loss with cosine distance \\
$\mathcal{L}_{\text{disp}}^{\text{hinge}}$ & Dispersive loss with hinge loss \\
$\mathcal{L}_{\text{diff}}$ & Standard diffusion training loss (abbreviated) \\
$\mathcal{L}_{\text{diffusion}}$ & Standard diffusion training loss \\
$\mathcal{L}_{\text{total}}$ & Combined loss (diffusion + dispersive) \\
$\mathcal{D}(h_i, h_j)$ & Distance function between representations $h_i$ and $h_j$ \\
$\tau$ & Temperature parameter for softmax/exponential functions \\
$\lambda$ & Weight coefficient for dispersive loss \\
$\epsilon$ & Margin parameter for hinge loss \\
\bottomrule
\end{tabular}
\caption{Notation and variable definitions for D²PPO framework - Part II}
\label{tab:notation_part2}
\end{table}

\subsubsection{From InfoNCE to Dispersive Loss}

The InfoNCE objective maximizes agreement between positive pairs $(z_i, z_i^+)$ while minimizing agreement with negative samples. The standard formulation can be written as:
\begin{align}
\mathcal{L}_{InfoNCE}  &= -\log \frac{\exp\left(-\frac{\mathcal{D}(z_i, z_i^+)}{\tau}\right)}{\exp\left(-\frac{\mathcal{D}(z_i, z_i^+)}{\tau}\right) + \sum_{j \neq i} \exp\left(-\frac{\mathcal{D}(z_i, z_j)}{\tau}\right)} \nonumber \\
&= -\log \frac{\exp\left(-\frac{\mathcal{D}(z_i, z_i^+)}{\tau}\right)}{\sum_{j=1}^{N} \exp\left(-\frac{\mathcal{D}(z_i, z_j)}{\tau}\right)}
\end{align}
where $\mathcal{D}(\cdot, \cdot)$ is a distance function, $\tau$ is the temperature parameter, and the denominator includes both positive and negative samples.

By applying logarithm properties, we can decompose InfoNCE into two distinct components:
\begin{align}
\mathcal{L}_{InfoNCE} &= \frac{\mathcal{D}(z_i, z_i^+)}{\tau} + \log \left( \exp\left(-\frac{\mathcal{D}(z_i, z_i^+)}{\tau}\right) + \sum_{j \neq i} \exp\left(-\frac{\mathcal{D}(z_i, z_j)}{\tau}\right) \right) \nonumber \\
&= \frac{\mathcal{D}(z_i, z_i^+)}{\tau} + \log \sum_{j=1}^{N} \exp\left(-\frac{\mathcal{D}(z_i, z_j)}{\tau}\right)
\end{align}
The first term encourages similarity between positive pairs, while the second term (log-sum-exp) acts as a normalization factor over all samples.

For the dispersive loss, we directly focus on the normalization term from InfoNCE which encourages dispersion among all samples. The key insight is to use only the log-sum-exp term without requiring positive pairs:

\begin{equation}
\mathcal{L}_{Disp} = \log \sum_{j=1}^{N} \exp\left(-\frac{\mathcal{D}(z_i, z_j)}{\tau}\right)
\end{equation}

To obtain a practical batch-level objective, we take the expectation over all sample indices in the batch:
\begin{equation}
\mathcal{L}_{Disp} = \mathbb{E}_{i} \left[ \log \sum_{j=1}^{B} \exp\left(-\frac{\mathcal{D}(z_i, z_j)}{\tau}\right) \right]
\end{equation}
This formulation encourages all representations in a batch to be maximally dispersed from each other, preventing representational collapse.

\subsubsection{Dispersive Loss Variants}

\textbf{InfoNCE L2 Variant:} Using squared L2 distance $\mathcal{D}(h_i, h_j) = ||h_i - h_j||_2^2$:
\begin{equation}
\mathcal{L}_{\text{disp}}^{\text{L2}} = \log \left( \frac{1}{B(B-1)} \sum_{i=1}^{B} \sum_{j \neq i} \exp\left(-\frac{||h_i - h_j||_2^2}{\tau}\right) \right)
\end{equation}

\textbf{InfoNCE Cosine Variant:} Using cosine dissimilarity $\mathcal{D}(h_i, h_j) = 1 - {h_i^T h_j}/{||h_i||_2 \cdot ||h_j||_2}$:
\begin{equation}
\mathcal{L}_{\text{disp}}^{\text{cos}} = \log \left( \frac{1}{B(B-1)} \sum_{i=1}^{B} \sum_{j \neq i} \exp\left(-\frac{1 - \frac{h_i^T h_j}{||h_i||_2 \cdot ||h_j||_2}}{\tau}\right) \right)
\end{equation}

\textbf{Hinge Loss Variant:} Enforces a minimum margin $\epsilon$ between representations:
\begin{equation}
\mathcal{L}_{\text{disp}}^{\text{hinge}} = \frac{1}{B(B-1)} \sum_{i=1}^{B} \sum_{j \neq i} \max(0, \epsilon - ||h_i - h_j||_2^2)^2
\end{equation}

\subsubsection{Mathematical Properties of Dispersive Loss}

\textbf{Gradient Analysis:} To understand how dispersive loss affects learning, we analyze its gradient with respect to individual representations. The gradient takes a weighted form:
\begin{equation}
\frac{\partial \mathcal{L}_{Disp}}{\partial h_i} = \frac{1}{\tau} \sum_{j=1}^B w_{ij} \frac{\partial \mathcal{D}(h_i, h_j)}{\partial h_i}
\end{equation}

where $w_{ij} = \frac{\exp\left(-\frac{\mathcal{D}(h_i, h_j)}{\tau}\right)}{\sum_{l=1}^B \exp\left(-\frac{\mathcal{D}(h_i, h_l)}{\tau}\right)}$ are attention weights that emphasize nearby representations for each anchor sample $i$.

For the commonly used L2 distance $\mathcal{D}(h_i, h_j) = ||h_i - h_j||_2^2$, the gradient becomes:
\begin{equation}
\frac{\partial \mathcal{L}_{Disp}}{\partial h_i} = \frac{2}{\tau} \sum_{j=1}^B w_{ij} (h_i - h_j)
\end{equation}

This gradient has an intuitive interpretation: it pushes representation $h_i$ away from other representations $h_j$, with the repulsion strength determined by their current similarity (through the attention weights $w_{ij}$). Closer representations experience stronger repulsive forces, naturally leading to dispersed representations.

\subsection{Policy Gradient Derivation for Diffusion Policies}

\subsubsection{Chain Rule Application}

For a diffusion policy, the action generation involves a denoising chain. The probability of generating action $a_t^0$ at environment timestep $t$ is:
\begin{equation}
\pi_\theta(a_t^0 | s_t) = \int p(a_t^K) \prod_{k=1}^{K} p_\theta(a_t^{k-1} | a_t^k, s_t) da_t^{1:K}
\end{equation}

Since $p(a_t^K)$ is independent of $\theta$, the policy gradient decomposes as:
\begin{equation}
\nabla_\theta \log \pi_\theta(a_t^0 | s_t) = \sum_{k=1}^{K} \nabla_\theta \log p_\theta(a_t^{k-1} | a_t^k, s_t)
\end{equation}

\subsubsection{Gaussian Denoising Steps}

In diffusion policies, each denoising step is modeled as a Gaussian distribution where the network predicts the mean, while the variance is fixed:
\begin{equation}
p_\theta(a_t^{k-1} | a_t^k, s_t) = \mathcal{N}(a_t^{k-1}; \mu_\theta(a_t^k, k, s_t), \sigma_k^2 I)
\end{equation}
For a Gaussian distribution $\mathcal{N}(x; \mu, \sigma^2)$, the log-probability is:
\begin{equation}
\log p(x) = -\frac{1}{2\sigma^2}(x - \mu)^2 + \text{const}
\end{equation}
Taking the gradient with respect to parameters $\theta$ that affect the mean $\mu_\theta$:
\begin{equation}
\nabla_\theta \log p_\theta(a_t^{k-1} | a_t^k, s_t) = \frac{1}{\sigma_k^2} (a_t^{k-1} - \mu_\theta(a_t^k, k, s_t)) \nabla_\theta \mu_\theta(a_t^k, k, s_t)
\end{equation}
This follows from the chain rule, where the derivative of $-\frac{1}{2\sigma^2}(x - \mu)^2$ with respect to $\mu$ is $\frac{1}{\sigma^2}(x - \mu)$.

To compute $\nabla_\theta \mu_\theta$, we need to understand how DDPM parameterizes the mean function. In standard DDPM, the predicted mean is defined as:
\begin{equation}
\mu_\theta(a_t^k, k, s_t) = \frac{1}{\sqrt{\alpha_k}} \left( a_t^k - \frac{\beta_k}{\sqrt{1-\bar{\alpha}_k}} \epsilon_\theta(a_t^k, k, s_t) \right)
\end{equation}
where $\epsilon_\theta$ is the neural network that predicts the noise added at timestep $k$.

Taking the gradient with respect to $\theta$:
\begin{align}
\nabla_\theta \mu_\theta &= \frac{1}{\sqrt{\alpha_k}} \nabla_\theta \left( a_t^k - \frac{\beta_k}{\sqrt{1-\bar{\alpha}_k}} \epsilon_\theta(a_t^k, k, s_t) \right) \nonumber \\
&= \frac{1}{\sqrt{\alpha_k}} \left( 0 - \frac{\beta_k}{\sqrt{1-\bar{\alpha}_k}} \nabla_\theta \epsilon_\theta(a_t^k, k, s_t) \right) \nonumber \\
&= -\frac{\beta_k}{\sqrt{\alpha_k(1-\bar{\alpha}_k)}} \nabla_\theta \epsilon_\theta(a_t^k, k, s_t)
\end{align}
Here, $\alpha_k$, $\beta_k$, and $\bar{\alpha}_k$ are the diffusion schedule parameters, with $\bar{\alpha}_k = \prod_{i=1}^k \alpha_i$ being the cumulative product.

\subsection{Efficient Gradient Estimation}

\subsubsection{Importance Sampling for Denoising Steps}

Computing gradients through all $K$ denoising steps is computationally expensive. We use importance sampling by selecting a subset $\mathcal{S} \subset \{1, 2, \ldots, K\}$.

The policy gradient requires summing over all $K$ denoising steps:
\begin{equation}
\nabla_\theta J(\theta) = \mathbb{E}_{\tau} \left[ \sum_{k=1}^{K} \nabla_\theta \log p_\theta(a_t^{k-1} | a_t^k, s_t) \cdot \hat{A}_t \right]
\end{equation}
where $\hat{A}_t$ is the advantage estimate for the complete action sequence.

Using importance sampling over a subset of timesteps, we obtain an unbiased estimate:
\begin{equation}
\nabla_\theta J(\theta) \approx \frac{1}{|\mathcal{S}|} \sum_{k \in \mathcal{S}} \frac{K}{p(k)} \nabla_\theta \log p_\theta(a_t^{k-1} | a_t^k, s_t) \cdot \hat{A}_t^{(k)}
\end{equation}
where $\mathcal{S}$ denotes the sampled subset of denoising steps, $|\mathcal{S}|$ is the subset size, $p(k)$ is the sampling probability for step $k$, and $\hat{A}_t^{(k)}$ is the step-conditioned advantage estimate. This formulation enables efficient policy gradient computation while maintaining the enhanced representational structure from pre-training.

\subsection{PPO Loss Adaptation for Diffusion Policies}

\subsubsection{PPO Objective Function}

Proximal Policy Optimization (PPO)~\citep{ppo2017} addresses the challenge of maintaining stable policy updates in reinforcement learning by constraining policy changes within a trust region. The key insight is to optimize a surrogate objective that approximates the policy improvement while preventing destructively large updates.

The standard PPO objective function with clipped probability ratios is:
\begin{equation}
\mathcal{L}^{PPO}(\theta) = \mathbb{E}_t \left[ \min(r_t(\theta) \hat{A}_t, \text{clip}(r_t(\theta), 1-\epsilon, 1+\epsilon) \hat{A}_t) \right]
\end{equation}
where $r_t(\theta) = \frac{\pi_\theta(a_t | s_t)}{\pi_{\theta_{old}}(a_t | s_t)}$ is the probability ratio between current and old policies, $\hat{A}_t$ is the advantage estimate, $\epsilon$ is the clipping parameter (typically 0.1-0.2), and $\text{clip}(x, a, b) = \max(a, \min(x, b))$ constrains $x$ to the interval $[a, b]$.

The clipping mechanism serves as a practical approximation to trust region constraints. When $\hat{A}_t > 0$ (good actions), the ratio is clipped at $1+\epsilon$ to prevent excessive policy updates. When $\hat{A}_t < 0$ (poor actions), the ratio is clipped at $1-\epsilon$ to limit policy degradation.

For diffusion policies, we adapt this objective by considering the multi-step denoising nature. The complete PPO loss for diffusion policies becomes:
\begin{equation}
\mathcal{L}^{D^2PPO}(\theta) = \mathbb{E}_t \left[ \sum_{k \in \mathcal{S}} \frac{K}{|\mathcal{S}|p(k)} \min(r_t^{(k)}(\theta) \hat{A}_t^{(k)}, \text{clip}(r_t^{(k)}(\theta), 1-\epsilon, 1+\epsilon) \hat{A}_t^{(k)}) \right]
\end{equation}
where the summation over sampled denoising steps $k \in \mathcal{S}$ accounts for the multi-step generation process in diffusion policies. The importance weighting $\frac{K}{|\mathcal{S}|p(k)}$ ensures unbiased gradient estimation when sampling subsets of denoising timesteps.

\subsubsection{Probability Ratio Computation}

PPO requires computing the probability ratio between the current policy $\pi_\theta$ and the old policy $\pi_{\theta_{old}}$ for importance sampling. For diffusion policies, this ratio is computed at each denoising timestep:
\begin{equation}
r_t^{(k)}(\theta) = \frac{p_\theta(a_t^{k-1} | a_t^k, s_t)}{p_{\theta_{old}}(a_t^{k-1} | a_t^k, s_t)}
\end{equation}

Since both policies use Gaussian distributions with identical variances, the ratio simplifies to:
\begin{equation}
r_t^{(k)}(\theta) = \exp\left( -\frac{1}{2\sigma_k^2} \left[ ||a_t^{k-1} - \mu_\theta||^2 - ||a_t^{k-1} - \mu_{\theta_{old}}||^2 \right] \right)
\end{equation}
This exponential form ensures numerical stability and efficient computation of the probability ratio.

\subsubsection{Advantage Estimation}

PPO requires accurate advantage estimation to guide policy updates. We employ Generalized Advantage Estimation (GAE)~\citep{gae2015} to compute advantages:
\begin{equation}
\hat{A}_t = \sum_{l=0}^{\infty} (\gamma \lambda)^l \delta_{t+l}
\end{equation}
where $\delta_t = r_t + \gamma V_\phi(s_{t+1}) - V_\phi(s_t)$ is the temporal difference error, $\gamma$ is the discount factor, and $\lambda$ is the GAE parameter that trades off bias and variance.

\section{\raggedright\Large Appendix C: Algorithm Pseudocode}

\algrenewcommand\algorithmicrequire{\textbf{Input:}}
\algrenewcommand\algorithmicensure{\textbf{Output:}}

\subsection{Pretraining with Dispersive Loss}

Algorithm~\ref{alg:pretraining} presents the complete pre-training procedure for D²PPO, which augments standard diffusion policy training with dispersive loss regularization to address representation collapse in diffusion-based robot control. The algorithm takes expert demonstrations and systematically combines diffusion loss with dispersive loss to prevent representation collapse while maintaining accurate noise prediction capabilities.

The pre-training stage serves as the foundation for enhanced representation learning. Standard diffusion policy training relies solely on reconstruction loss, which optimizes for denoising accuracy but neglects feature diversity, leading to similar observations being mapped to indistinguishable representations. Our enhanced pre-training addresses this limitation by introducing dispersive regularization that encourages representational diversity without requiring positive pair construction.

Key algorithmic features include: (1) unified objective function $\mathcal{L}_{\text{total}} = \mathcal{L}_{\text{diff}} + \lambda \mathcal{L}_{\text{disp}}$ that balances denoising accuracy with representation diversity, (2) flexible dispersive loss variants (InfoNCE-L2, InfoNCE-Cosine, Hinge) enabling task-specific optimization strategies, (3) strategic intermediate feature extraction through network hooks that target specific MLP layers for regularization, (4) batch-wise dispersive loss computation ensuring representational diversity within each training batch, and (5) temperature-controlled dispersion strength that provides fine-grained control over regularization intensity. This enhanced pre-training establishes a robust representational foundation that significantly improves subsequent policy gradient optimization effectiveness.

\begin{algorithm}[H]
\caption{Diffusion Policy Pretraining with Dispersive Loss}
\label{alg:pretraining}
\begin{algorithmic}[1]
\Require Expert demonstrations $\mathcal{D} = \{(s_i, a_i)\}_{i=1}^N$
\Require Diffusion model $\epsilon_\theta$, dispersive loss weight $\lambda$, temperature $\tau$
\Ensure Pretrained diffusion policy $\epsilon_{\theta^*}$

\State Initialize model parameters $\theta$
\For{epoch $= 1, 2, \ldots, E_{pre}$}
    \For{batch $\mathcal{B} \subset \mathcal{D}$}
        \State Sample timesteps $k \sim \mathcal{U}(1, K)$ for each sample in $\mathcal{B}$
        \State Sample noise $\epsilon \sim \mathcal{N}(0, I)$
        
        \State \textbf{// Standard Diffusion Loss}
        \State $a^k = \sqrt{\bar{\alpha}_k} a^0 + \sqrt{1-\bar{\alpha}_k} \epsilon$
        \State $\hat{\epsilon} = \epsilon_\theta(a^k, s, k)$
        \State $\mathcal{L}_{diff} = \|\epsilon - \hat{\epsilon}\|_2^2$
        
        \State \textbf{// Dispersive Loss Computation}
        \State Extract intermediate representations $H = \{h_i\}_{i=1}^B$ using network hooks
        \If{$|B| > 1$}
            \If{dispersive\_type == "infonce\_l2"}
                \State Compute pairwise L2 distances: $D_{ij} = \|h_i - h_j\|_2^2$
                \State $\mathcal{L}_{\text{disp}} = \log \left( \frac{1}{B^2} \sum_{i,j} \exp(-D_{ij} / \tau) \right)$
            \ElsIf{dispersive\_type == "infonce\_cosine"}
                \State Normalize: $\hat{h}_i = h_i / \|h_i\|_2$
                \State Compute cosine dissimilarities: $D_{ij} = 1 - \hat{h}_i \cdot \hat{h}_j$
                \State $\mathcal{L}_{\text{disp}} = \log \left( \frac{1}{B^2} \sum_{i,j} \exp(-D_{ij} / \tau) \right)$
            \ElsIf{dispersive\_type == "hinge"}
                \State $\mathcal{L}_{\text{disp}} = \frac{1}{B^2} \sum_{i,j} \max(0, \epsilon - \|h_i - h_j\|_2^2)^2$
            \EndIf
        \Else
            \State $\mathcal{L}_{\text{disp}} = 0$
        \EndIf
        
        \State \textbf{// Combined Loss and Update}
        \State $\mathcal{L}_{\text{total}} = \mathcal{L}_{\text{diff}} + \lambda \cdot \mathcal{L}_{\text{disp}}$
        \State Update parameters: $\theta \leftarrow \theta - \alpha \nabla_{\theta} \mathcal{L}_{\text{total}}$
    \EndFor
\EndFor
\State \Return $\epsilon_{\theta^*}$
\end{algorithmic}
\end{algorithm}

\subsection{PPO Fine-tuning for Diffusion Policies}

Algorithm~\ref{alg:finetuning} details the policy gradient fine-tuning stage that leverages the enhanced representations from pre-training to optimize task performance through reinforcement learning. This algorithm adapts standard PPO to the multi-step denoising nature of diffusion policies, enabling direct task-oriented optimization while preserving the representational diversity established during pre-training.

The fine-tuning stage transforms the pre-trained diffusion policy into a task-specific controller optimized for success rate maximization. Unlike standard policy gradient methods that operate on single-step action generation, this algorithm must handle the inherent complexity of diffusion policies where actions are generated through iterative denoising processes. The challenge lies in computing gradients through the entire denoising chain while maintaining computational efficiency and training stability.

Core algorithmic components include: (1) iterative environment interaction using the complete $K$-step denoising chain for action generation, where each trajectory involves multiple forward passes through the diffusion model, (2) GAE-based advantage estimation with temporal difference learning that provides stable policy gradient signals, (3) PPO clipped objective function specifically adapted for diffusion policies with importance sampling across denoising timesteps, (4) joint policy and value function optimization that balances exploration and exploitation, (5) proper handling of probability ratios across denoising timesteps using chain rule decomposition to maintain training stability, and (6) strategic timestep sampling for computational efficiency without sacrificing gradient quality. This comprehensive fine-tuning approach ensures that the enhanced representations from pre-training translate into superior task performance while maintaining the representational structure that prevents collapse.

\begin{algorithm}[H]
\caption{PPO Fine-tuning for Diffusion Policies}
\label{alg:finetuning}
\begin{algorithmic}[1]
\Require Pretrained model $\epsilon_{\theta_{old}}$, environment $\mathcal{E}$, value network $V_\phi$
\Require PPO hyperparameters: $\epsilon_{clip}$, GAE parameter $\lambda_{gae}$
\Ensure Fine-tuned policy $\pi_{\theta}$

\State Initialize policy parameters $\theta \leftarrow \theta_{old}$, value parameters $\phi$

\For{iteration $= 1, 2, \ldots, M$}
    \State \textbf{// Data Collection Phase}
    \State Initialize trajectory buffer $\mathcal{T} = \{\}$
    \For{each environment episode}
        \State Reset environment: $s_0 \sim \rho_0$
        \For{$t = 0, 1, \ldots, T-1$}
            \State Sample noise $a_t^K \sim \mathcal{N}(0, I)$
            \For{$k = K, K-1, \ldots, 1$} \Comment{Denoising process}
                \State $a_t^{k-1} = \text{DenoisingStep}(a_t^k, \epsilon_\theta(a_t^k, s_t, k), k)$
            \EndFor
            \State Execute action $a_t = a_t^0$, observe $(s_{t+1}, r_t)$
            \State Store $(s_t, a_t, r_t, \log \pi_\theta(a_t^0|s_t))$ in $\mathcal{T}$
        \EndFor
    \EndFor
    
    \State \textbf{// Advantage Estimation}
    \State Compute returns $R_t$ and GAE advantages $\hat{A}_t$ using $V_\phi$
    
    \State \textbf{// PPO Update}
    \For{PPO epoch $= 1, 2, \ldots, K_{ppo}$}
        \For{mini-batch $\mathcal{B} \subset \mathcal{T}$}
            \State Compute current log-probabilities $\log \pi_\theta(a_t^0|s_t)$
            \State $r_t^{(k)}(\theta) = \exp(\log \pi_\theta(a_t^0|s_t) - \log \pi_{old}(a_t^0|s_t))$
            \State $\mathcal{L}_{clip} = -\min(r_t^{(k)} \hat{A}_t, \text{clip}(r_t^{(k)}, 1-\epsilon_{clip}, 1+\epsilon_{clip}) \hat{A}_t)$
            \State $\mathcal{L}_{value} = (V_\phi(s_t) - R_t)^2$
            \State $\mathcal{L}_{PPO} = \mathcal{L}_{clip} + c_1 \mathcal{L}_{value}$
            \State Update $\theta, \phi$ using $\nabla \mathcal{L}_{PPO}$
        \EndFor
    \EndFor
    \State Update old policy: $\theta_{old} \leftarrow \theta$
\EndFor

\State \Return Fine-tuned policy $\pi_{\theta}$
\end{algorithmic}
\end{algorithm}

\section{\raggedright\Large Appendix D: Implementation Details and Experimental Configuration}

\subsection{D²PPO Training Framework}

Our D²PPO approach employs a two-stage training framework that integrates dispersive regularization with standard diffusion policy learning. The MLP backbone architecture is enhanced with strategic feature extraction capabilities through forward hooks registered at key network layers without modifying the core structure. We systematically evaluate three layer placement options: early layers (first hidden layer) for capturing low-level representations crucial for basic manipulation tasks, mid-layers (intermediate hidden layers) where high-level semantic features concentrate, and late layers (before final output) for refined representations encoding complex behavioral patterns.

During pre-training, intermediate representations are extracted from selected MLP layers and processed through global average pooling to ensure consistent dimensionality. The training combines standard diffusion loss with dispersive regularization as $\mathcal{L}_{\text{total}} = \mathcal{L}_{\text{diffusion}} + \lambda \mathcal{L}_{\text{dispersive}}$, where gradients from both components flow through shared parameters to encourage representational diversity. The fine-tuning stage follows the DPPO framework with nested MDP formulation, sampling subsets of denoising timesteps for computational efficiency while computing policy gradients through backpropagation with proper weighting. Standard PPO loss integration with clipping and advantage estimation is carefully adapted for diffusion policy parameterization.

\subsection{Dataset Analysis and Task Characteristics}

Our experimental evaluation is conducted on the RoboMimic benchmark, which provides a comprehensive suite of robotic manipulation tasks with varying complexity levels. The dataset analysis reveals important insights that inform our understanding of task difficulty and the corresponding effectiveness of dispersive loss regularization.

\begin{figure}[ht]
\centering
\includegraphics[width=\textwidth]{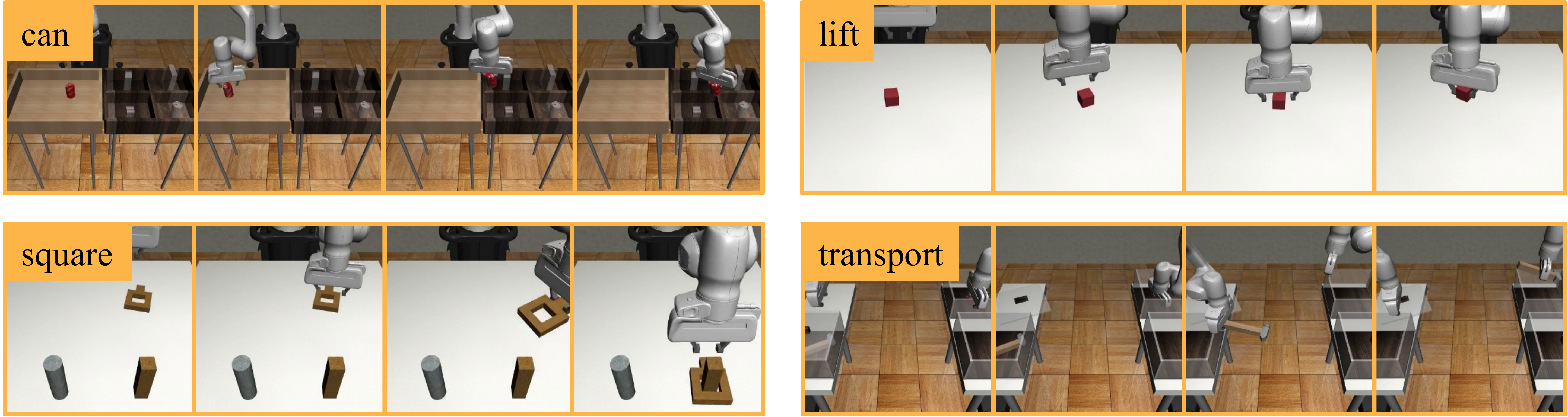}
\caption{Four robotic manipulation task scenarios in the RoboMimic benchmark. This figure illustrates the core tasks used to evaluate robot policy learning and generalization capabilities: \textbf{(Top Left) Can Task}: Grasping a red can from the table and moving it to a designated position, emphasizing grasping angle flexibility; \textbf{(Top Right) Lift Task}: Picking up a cube from the table surface and lifting it away, suitable for testing basic grasping actions; \textbf{(Bottom Left) Square Task}: Inserting a square block into the corresponding square hole, emphasizing fine alignment and high-precision operation capabilities; \textbf{(Bottom Right) Transport Task}: Using dual-arm coordination to transport objects, representing the most complex task that emphasizes dual-arm coordination, long-term planning, and multi-stage operations. These four tasks represent different levels of task complexity and motion precision requirements, ranging from basic motion control (Lift, Can) to moderate precision challenges (Square) to complex motion coordination and planning (Transport).}
\label{fig:robomimic_tasks}
\end{figure}

\begin{figure}[ht]
\centering
\includegraphics[width=\textwidth]{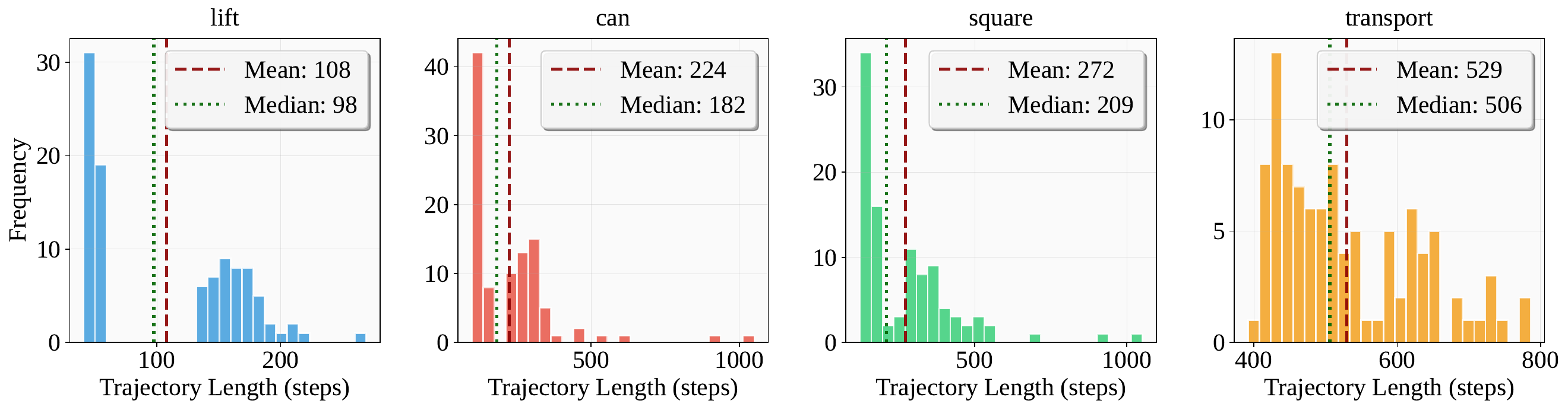}
\caption{Distribution analysis of trajectory lengths across four RoboMimic tasks. The histogram shows a clear task complexity hierarchy with Transport requiring the longest execution sequences, followed by Square, Can, and Lift tasks. The varying distributions indicate different solution strategy requirements across tasks.}
\label{fig:dataset_analysis}
\end{figure}

The dataset characteristics provide crucial insights into task complexity and expert demonstration quality. The benchmark consists of 400 expert demonstration trajectories with 100 trajectories per task, ensuring balanced representation across all manipulation scenarios, where all trajectories represent successful task completions from expert demonstrations, providing high-quality learning targets for imitation learning. Analysis of trajectory lengths reveals a clear complexity hierarchy: Transport (529 average steps) > Square (272 steps) > Can (224 steps) > Lift (108 steps), which correlates strongly with our empirical findings that more complex tasks benefit increasingly from dispersive loss regularization, as shown in Figure~\ref{fig:dataset_analysis}. 

Square and Can tasks exhibit higher standard deviations in trajectory lengths, indicating more diverse solution strategies and higher behavioral complexity, suggesting that these tasks require more nuanced policy representations and explaining why dispersive loss provides substantial improvements for these scenarios. Each task presents distinct challenges: Lift requires basic object manipulation with minimal spatial precision; Can involves cylindrical object grasping with moderate coordination requirements; Square demands precise peg-in-hole insertion with tight spatial tolerances; Transport necessitates complex multi-object coordination and spatial reasoning across extended sequences.

\begin{figure}[ht]
\centering
\includegraphics[width=\textwidth]{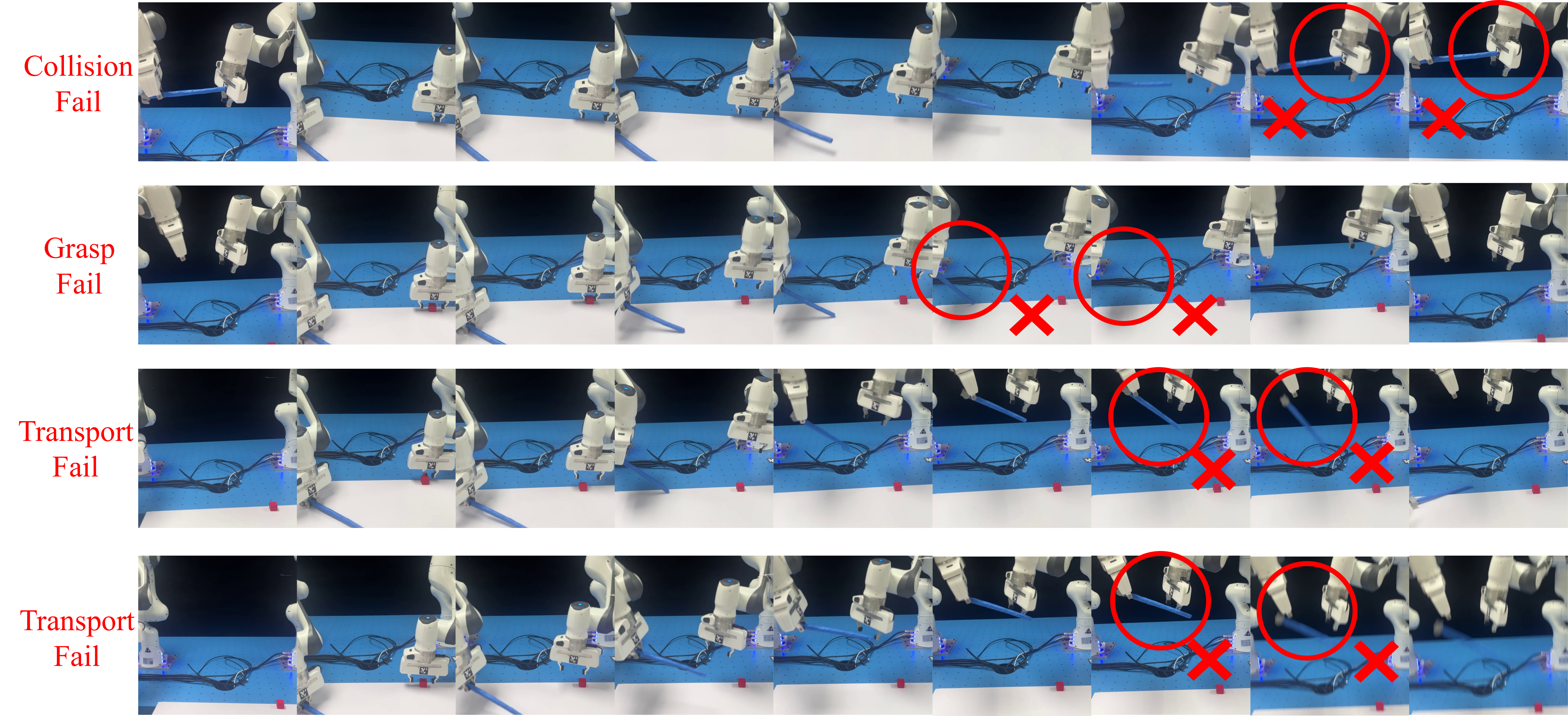}
\caption{Representative failure cases in the Transport task without dispersive regularization. Top to bottom: (1) Collision failure due to direct contact between arms during approach; (2) Post-grasp collision caused by the left arm's motion path interfering with the right arm; (3) Transport failure caused by premature release before the receiving arm grasps the object; (4) Transport failure due to inaccurate delivery, where the left arm fails to align the object with the right gripper's center. These failure modes highlight the importance of precise spatial-temporal coordination that dispersive loss regularization helps achieve.}
\label{fig:transport_failures}
\end{figure}

\begin{figure}[H]
\centering
\includegraphics[width=0.85\textwidth]{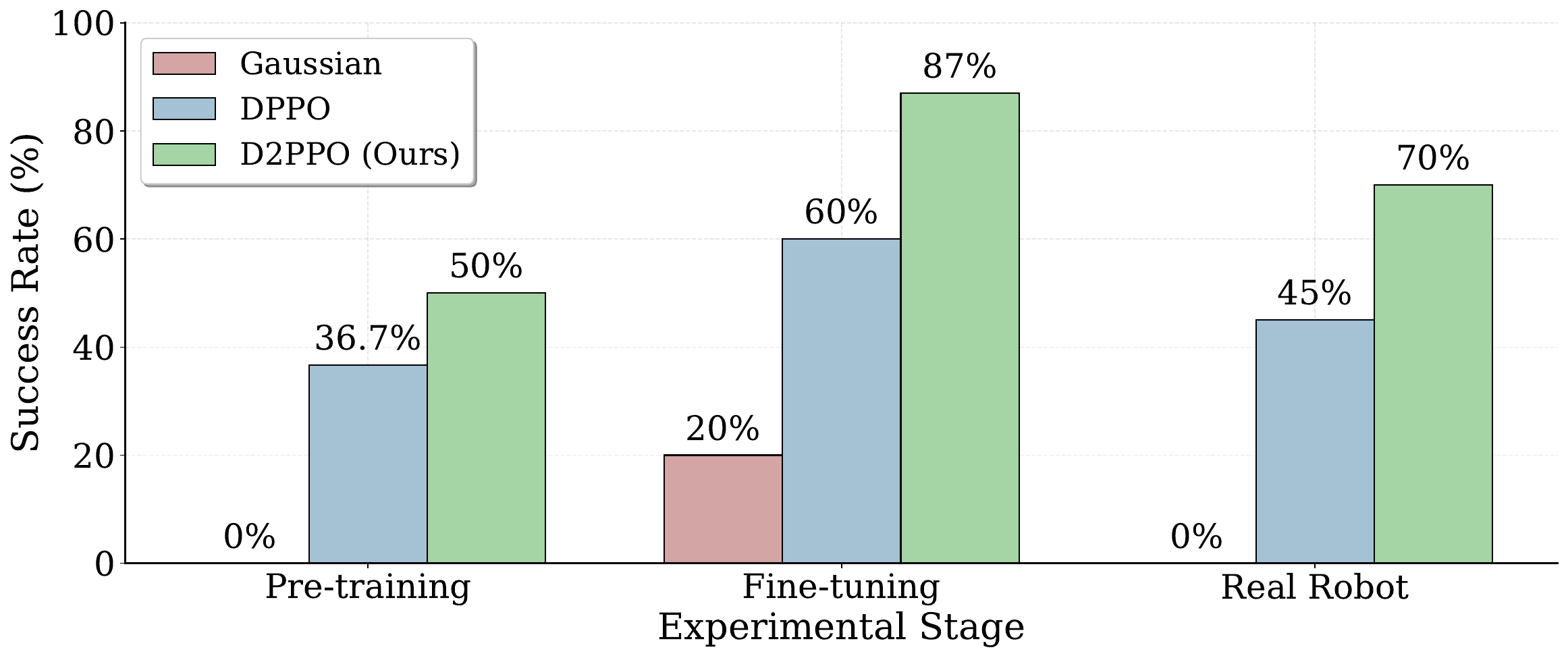}
\caption{Transport Task Performance Across Three Experimental Stages. Performance comparison showing success rates for Gaussian (muted red), DPPO (muted blue), and D²PPO (muted green) across pre-training, fine-tuning, and real robot experimental stages. Pre-training results show progressive improvement: Gaussian (0\%), DPPO (36.7\%), D²PPO (50\%). Fine-tuning demonstrates larger gaps: Gaussian (20\%), DPPO (60\%), D²PPO (87\%). Real robot deployment confirms practical superiority: Gaussian (0\%), DPPO (45\%), D²PPO (70\%).}
\label{fig:transport_task_stages}
\end{figure}

\subsection{Additional Experimental Results}

Figure~\ref{fig:transport_failures} illustrates the four primary failure categories observed in Transport tasks when dispersive regularization is absent, providing crucial insights into why dispersive loss regularization is particularly effective for complex manipulation scenarios. The collision-based failures (Rows 1-2) demonstrate how representation collapse leads to insufficient spatial awareness between dual arms, where similar spatial configurations are mapped to nearly identical feature representations, preventing the policy from distinguishing between safe and collision-prone approach trajectories and highlighting the critical need for diverse representations that can capture subtle spatial relationships. The transport failures (Rows 3-4) reveal timing and precision deficits that stem from collapsed representations, where the premature release failure indicates poor temporal coordination with similar visual states leading to identical timing decisions, and the delivery alignment failure demonstrates insufficient spatial precision where collapsed representations cannot distinguish between accurate and inaccurate object positioning relative to the receiving gripper.

These failure modes directly validate our hypothesis that representation quality becomes increasingly critical with task complexity, as the Transport task's requirement for precise dual-arm coordination, accurate object handoffs, and fine spatial-temporal alignment makes it particularly sensitive to representation collapse. By encouraging feature diversity through dispersive regularization, D²PPO enables the policy to maintain distinct representations for similar but critically different states, thereby preventing these systematic failure patterns. These dataset insights directly inform our experimental design and provide theoretical justification for the observed correlation between task complexity and dispersive loss effectiveness, validating our approach of using log-normalized execution steps as a complexity metric.

Figure~\ref{fig:transport_task_stages} provides a comprehensive view of Transport task performance evolution across the complete experimental pipeline. The results illustrate several key insights: (1) \textbf{Progressive Learning Benefits:} Pre-training establishes foundational advantages with D²PPO (50\%) outperforming DPPO (36.7\%) and Gaussian (0\%), demonstrating early benefits of dispersive regularization. (2) \textbf{Amplified Fine-tuning Gains:} Fine-tuning significantly amplifies performance differences, with D²PPO achieving 87\% success rate compared to DPPO's 60\% and Gaussian's 20\%, validating the synergistic effect of enhanced representations and policy gradient optimization. (3) \textbf{Robust Real-World Transfer:} Real robot deployment confirms practical superiority with D²PPO (70\%) substantially outperforming DPPO (45\%) and Gaussian (0\%), while maintaining only a 17\% sim-to-real gap compared to DPPO's 25\% gap, highlighting superior generalization capabilities.

\subsection{Hyperparameter Configuration and Selection}

Our experimental framework focuses on two critical hyperparameters that most significantly impact D²PPO performance: **dispersive loss coefficient ($\lambda$)** and **layer placement**.

\textbf{Dispersive Loss Coefficient Analysis:} We conduct detailed ablation studies specifically on the Square task, which serves as a representative high-precision manipulation scenario. As demonstrated in Figure~\ref{fig:lambda_effect_square}, systematic evaluation across $\lambda \in \{0.1, 0.3, 0.5, 0.7, 0.9\}$ reveals a non-monotonic relationship with performance. The optimal value $\lambda = 0.5$ achieves 14.3\% improvement over baseline, while insufficient regularization ($\lambda = 0.3$) degrades below baseline performance. This indicates that balanced dispersive regularization is crucial—too little fails to prevent representation collapse, while excessive values may interfere with task-relevant learning.

\textbf{Layer Placement Analysis:} We systematically evaluate dispersive loss effectiveness across different network layers through five variants: InfoNCE L2 Early/Mid/Late, InfoNCE Cosine Early, and Hinge Early. Figure~\ref{fig:comprehensive_four_subplots}(d) provides empirical guidance through a method suitability matrix heatmap, establishing that simple tasks achieve optimal performance with early-layer application while complex tasks require late-layer regularization. This layer-dependent effectiveness reflects the hierarchical nature of learned representations, where early layers capture fundamental motion primitives and late layers encode sophisticated spatial-temporal relationships.

\begin{table}[ht]
\centering
\small
\begin{tabular}{lcccc}
\toprule
\textbf{Parameter} & \textbf{Can Task} & \textbf{Lift Task} & \textbf{Square Task} & \textbf{Transport Task} \\
\midrule
\multicolumn{5}{c}{\textit{Diffusion Model Pre-training Configuration}} \\
Training Iterations & 8000 & 8000 & 8000 & 8000 \\
Batch Size & 256 & 256 & 256 & 256 \\
Learning Rate & $1 \times 10^{-4}$ & $1 \times 10^{-4}$ & $1 \times 10^{-4}$ & $1 \times 10^{-4}$ \\
Dispersive Loss Weight ($\lambda$) & 0.5 & 0.5 & 0.5 & 0.5 \\
Temperature ($\tau$) & 0.5 & 0.5 & 0.5 & 0.5 \\
Layer Placement & Early & Early & Early & Late \\
Denoising Steps & 100 & 100 & 100 & 100 \\
Horizon Steps & 4 & 4 & 4 & 8 \\
Weight Decay & 0 & 0 & 0 & 0 \\
Optimizer & AdamW & AdamW & AdamW & AdamW \\
\midrule
\multicolumn{5}{c}{\textit{Task-Specific Environment Parameters}} \\
Observation Dimension & 9 & 9 & 9 & 18 \\
Action Dimension & 7 & 7 & 7 & 14 \\
Image Resolution & $96 \times 96$ & $96 \times 96$ & $96 \times 96$ & $96 \times 96$ \\
Number of Cameras & 1 & 1 & 1 & 2 \\
Dataset Size & 100 trajectories & 100 trajectories & 100 trajectories & 100 trajectories \\
\bottomrule
\end{tabular}
\caption{Pre-training hyperparameter specifications for diffusion policy learning across all experimental tasks.}
\label{tab:pretraining_hyperparameters}
\end{table}

\subsubsection{Network Architecture Specifications}

All tasks employ a unified Vision Diffusion MLP architecture optimized for robotic manipulation, featuring a Vision Transformer (ViT) encoder backbone with patch size 8, depth 1, and embedding dimension 128. The architecture incorporates MLP dimensions of [768, 768, 768] with residual connections for enhanced gradient flow, 128-dimensional spatial embeddings for encoding spatial relationships, and 32-dimensional time embeddings for temporal context in the diffusion process. Data augmentation is enabled during training to improve generalization across visual variations. The architecture design balances representational capacity with computational efficiency, utilizing attention mechanisms for spatial reasoning while maintaining fast inference speeds required for real-time robotic control.

\subsubsection{Fine-tuning Configuration}

The fine-tuning phase employs PPO-based reinforcement learning with task-specific optimizations and diffusion-specific parameters. Table~\ref{tab:finetuning_hyperparameters} provides comprehensive specifications for all fine-tuning hyperparameters.

\begin{table}[ht]
\centering
\small
\begin{tabular}{lcccc}
\toprule
\textbf{Parameter} & \textbf{Can Task} & \textbf{Lift Task} & \textbf{Square Task} & \textbf{Transport Task} \\
\midrule
\multicolumn{5}{c}{\textit{PPO Training Configuration}} \\
Total Iterations & 300 & 300 & 300 & 1000 \\
Steps per Iteration & 300 & 300 & 300 & 400 \\
Total Training Steps & 90,000 & 90,000 & 90,000 & 400,000 \\
Batch Size & 7,500 & 7,500 & 7,500 & 10,000 \\
Actor Learning Rate & $1 \times 10^{-5}$ & $1 \times 10^{-5}$ & $1 \times 10^{-5}$ & $1 \times 10^{-5}$ \\
Critic Learning Rate & $1 \times 10^{-3}$ & $1 \times 10^{-3}$ & $1 \times 10^{-3}$ & $1 \times 10^{-3}$ \\
Discount Factor ($\gamma$) & 0.999 & 0.999 & 0.999 & 0.999 \\
GAE Lambda & 0.95 & 0.95 & 0.95 & 0.95 \\
\midrule
\multicolumn{5}{c}{\textit{PPO Algorithm Parameters}} \\
Update Epochs & 10 & 10 & 10 & 8 \\
Value Function Coefficient & 0.5 & 0.5 & 0.5 & 0.5 \\
Target KL Divergence & 1.0 & 1.0 & 1.0 & 1.0 \\
PPO Clip Coefficient & 0.01 & 0.01 & 0.01 & 0.001 \\
PPO Clip Base & 0.001 & 0.001 & 0.001 & 0.0001 \\
\midrule
\multicolumn{5}{c}{\textit{Environment and Execution Configuration}} \\
Parallel Environments & 50 & 50 & 50 & 50 \\
Max Episode Steps & 300 & 300 & 400 & 700 \\
Action Steps per Policy Call & 4 & 4 & 4 & 8 \\
Fine-tuning Denoising Steps & 10 & 10 & 10 & 10 \\
Min Sampling Std & 0.1 & 0.1 & 0.1 & 0.08 \\
Observation Normalization & True & True & True & True \\
\midrule
\multicolumn{5}{c}{\textit{Monitoring and Evaluation}} \\
Validation Frequency & Every 10 iterations & Every 10 iterations & Every 10 iterations & Every 10 iterations \\
Save Model Frequency & Every 100 iterations & Every 100 iterations & Every 100 iterations & Every 100 iterations \\
WandB Logging & Enabled & Enabled & Enabled & Enabled \\
\bottomrule
\end{tabular}
\caption{Fine-tuning hyperparameter specifications for PPO-based policy optimization across all experimental tasks.}
\label{tab:finetuning_hyperparameters}
\end{table}

\subsection{Experimental Infrastructure and Computational Setup}

Our experimental framework employs a comprehensive infrastructure designed for reproducible and scalable robotic manipulation research. The system utilizes Hydra configuration management with hierarchical organization across pre-training, fine-tuning, and evaluation phases, supporting command-line override for efficient hyperparameter sweeps and ablation studies. All experiments are conducted on a high-performance computing server equipped with dual NVIDIA GeForce RTX 4090 GPUs (24 GB GDDR6X memory each), Intel Xeon Silver 4314 CPU (32 cores @ 2.40GHz), and 256 GB DDR4 RAM running Ubuntu 20.04.1 LTS with Linux kernel 5.15.0-139-generic. The software environment includes Python 3.8, PyTorch 2.4.0, CUDA 12.2, and key dependencies including Hydra Core 1.3.2, Robomimic, Robosuite v1.4.1, and MuJoCo 2.1.0. To ensure reproducible results, comprehensive random seed management is implemented throughout the codebase with a default seed value of 42, set consistently across Python's random module, NumPy, and PyTorch for both CPU and GPU operations.

\subsection{Performance Analysis and Task Complexity Relationships}

Our comprehensive evaluation across multiple task complexities reveals three critical insights. First, universal effectiveness is demonstrated through positive improvements across almost all manipulation tasks, validating broad applicability throughout the robotic manipulation spectrum. Second, complexity correlation analysis shows that more complex tasks (Transport, Square) consistently exhibit larger improvements, providing empirical evidence that representation quality becomes increasingly critical as task complexity increases. Third, method specialization patterns emerge where early-layer methods systematically excel for simpler tasks while late-layer methods dominate complex scenarios, suggesting that different levels of representation abstraction are optimal for different manipulation complexities.

Task-specific performance patterns validate the complexity-effectiveness relationship through distinct behavioral signatures. The Lift task demonstrates high baseline performance (87.0\%), creating a ceiling effect where dispersive methods achieve modest but consistent improvements (+3.4\% maximum). The Can task shows significant improvement potential with dispersive loss, where InfoNCE-L2 Early achieves a 73.3\% success rate compared to the 61.2\% baseline (+19.8\% improvement). The Square and Transport tasks exhibit the most substantial improvement potential, with InfoNCE-L2 Late achieving the largest relative improvements (+31.3\% and +36.2\% respectively), confirming that complex manipulation tasks benefit most significantly from enhanced representation learning.

\subsection{Method Variants and Layer Selection Analysis}

Layer selection experiments reveal systematic patterns in optimal dispersive loss application. For simple tasks like Lift, early layers (first quarter of the MLP) provide optimal results, achieving +3.4\% improvement. This finding suggests that basic representation diversity established early in network architecture suffices for straightforward manipulation tasks requiring primarily low-level motor control. In contrast, complex tasks like Transport benefit most from late layer regularization (final quarter of the network), achieving +36.2\% improvement. This pattern indicates that complex tasks require sophisticated high-level feature diversity emerging in deeper network layers, where abstract spatial-temporal relationships are encoded.

Among dispersive loss variants, InfoNCE-based loss with L2 distance demonstrates the most consistent performance across all task complexities. The cosine similarity variant shows more variable performance characteristics, particularly excelling in complex Transport tasks (+22.6\%) due to scale-invariant properties, but underperforming in simpler Lift tasks (-5.7\%) where absolute feature magnitudes may be more informative. The hinge loss variant provides competitive results for simpler manipulation tasks, matching InfoNCE-L2 Early performance in Lift scenarios, but shows systematically diminished effectiveness for complex manipulation scenarios requiring nuanced feature discrimination.

\end{document}